\documentclass[10pt,twocolumn,twoside] {IEEEtran}

\usepackage{threeparttable}
\usepackage{times}
\usepackage{epsfig}
\usepackage{graphicx}
\usepackage{amsmath}
\usepackage{amssymb}

\usepackage{subfigure}
\usepackage[section]{algorithm}
\usepackage{algorithmic}

\usepackage{makecell}
\usepackage{multirow}
\usepackage{array}
\usepackage{colortbl}
\usepackage{tikz}
\usepackage{enumitem}
\usepackage{comment}

\newcolumntype{L}[1]{>{\raggedright\let\newline\\\arraybackslash\hspace{0pt}}m{#1}}
\newcolumntype{C}[1]{>{\centering\let\newline\\\arraybackslash\hspace{0pt}}m{#1}}
\newcolumntype{R}[1]{>{\raggedleft\let\newline\\\arraybackslash\hspace{0pt}}m{#1}}

\newcommand{\eg}{e.g.}
\newcommand{\ie}{i.e.}

\newcommand{\cL}{\mathcal{L}}

\title{Learning Non-local Image Diffusion for\\ Image Denoising}
\author{Peng~Qiao, Yong~Dou, Wensen~Feng and Yunjin~Chen
\thanks{
P.~Qiao, Y.~Dou and Y.~Chen are with the
National Laboratory for Parallel and Distributed Processing,
School of Computer,
National University of Defense Technology, Changsha, 410073, China.
e-mail: (\{pengqiao, yongdou\}@nudt.edu.cn, \{chenyunjin\_nudt\}@hotmail.com).

W.~Feng is with
School of Automation and Electrical Engineering,
University of Science and Technology Beijing, Beijing.
e-mail: sanmumuren@ustb.edu.cn.
}
\thanks{This work was supported by the
   National Natural Science Foundation of China under Grant U1435219, 61402507, 61303070}}

\begin{document}

\markboth{IEEE TRANSACTIONS ON IMAGE PROCESSING,~Vol.~xx, No.~xx, 20xx}
{A \MakeLowercase{\textit{et al.}}: Learning Non-local Reaction Diffusion for Image Denoising}
\maketitle

\begin{abstract}
Image diffusion plays a fundamental role for the task of image denoising. Recently proposed trainable nonlinear reaction diffusion (TNRD) model defines a simple but very effective framework for image denoising. However, as the TNRD model is a local model, the diffusion behavior of which is purely controlled by information of local patches, it is prone to create artifacts in the homogenous regions and over-smooth highly textured regions, especially in the case of strong noise levels. Meanwhile, it is widely known that the non-local self-similarity (NSS) prior stands as an effective image prior for image denoising, which has been widely exploited in many non-local methods. In this work, we are highly motivated to embed the NSS prior into the TNRD model to tackle its weaknesses. In order to preserve the expected property that end-to-end training is available, we exploit the NSS prior by a set of non-local filters, and derive our proposed trainable non-local reaction diffusion (TNLRD) model for image denoising. Together with the local filters and influence functions, the non-local filters are learned by employing loss-specific training. The experimental results show that the trained  TNLRD model produces visually plausible recovered images with more textures and less artifacts, compared to its local versions. Moreover, the trained TNLRD model can achieve strongly competitive performance to recent state-of-the-art image denoising methods in terms of peak signal-to-noise ratio (PSNR) and structural similarity index (SSIM).
\end{abstract}

\begin{IEEEkeywords}
image denoising, non-local self-similarity, reaction diffusion, loss-based training.
\end{IEEEkeywords}

\vspace*{-0.25cm}
\section{Introduction}
\IEEEPARstart{I}{mage} denoising is one of the most fundamental processing
in image processing and low-level computer vision.
While it has been extensively studied, image denoising is still
an active topic in image processing and computer vision.
The goal of image denoising is to recover the clean image $u$ from
its noisy observation $f$, which is formulated as
\begin{equation}\label{denoiseProblem}
f = u + v,
\end{equation}
where $v$ is the noise.
In this paper, we assume $v$ is the additive Gaussian noise
with zero mean and standard derivation $\sigma$.

During the past decades, a large number of new image denoising methods
are continuously emerging. It is a difficult task to precisely categorize
existing image denoising approaches. Generally speaking,
most image denoising
approaches can be categorized as spatial domain and transform domain based
methods. Transform domain based methods first
represent an image with certain orthonormal transform,
such as wavelets \cite{simoncelli1996noise},
curvelets \cite{starck2002curvelet},
contourlets \cite{do2005contourlet},
or bandelets \cite{le2005sparse}, and
then attempt to separate noise from the clean image by manipulating the
coefficients according to the statistical characteristics of the clean image
and noise.

Spatial domain based approaches attempt to utilize the correlations
between adjacent pixels in an image. Depending on the way how to select
those adjacent pixels, spatial domain based methods can be categorized as
local and non-local methods.
In local methods, only those adjacent pixels in a spatial neighborhood
(probably with fixed shape and size) of the test pixel are investigated.
Pixels in this small spatial range are named as an image patch,
and the clean pixel value is estimated from this local patch.
A large number of local algorithms have been proposed, including
filtering based methods \cite{tomasi1998bilateral, elad2002origin,
takeda2007kernel, haykin2003least}, anisotropic diffusion based
methods \cite{perona1990scale, weickert1998anisotropic,
gilboa2002forward}, variational methods with various image
regularizers \cite{rudin1992nonlinear, chan2000high, roth2005fields},
and patch-based models via sparse representation
\cite{elad2006image, aharon2006k}.

\begin{figure}[t!]
\centering
\subfigure[]
{\includegraphics[width=0.6\linewidth]{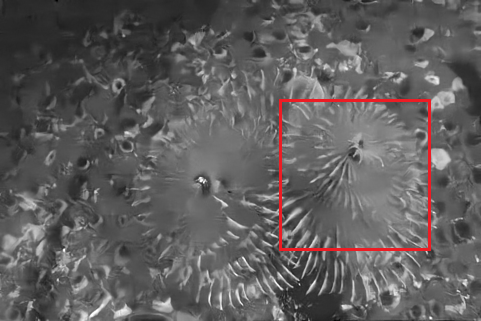}}\hfill
\subfigure[]
{\includegraphics[width=0.36\linewidth]{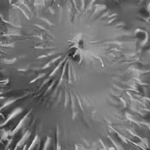}}\\
\subfigure[]
{\includegraphics[width=0.6\linewidth]{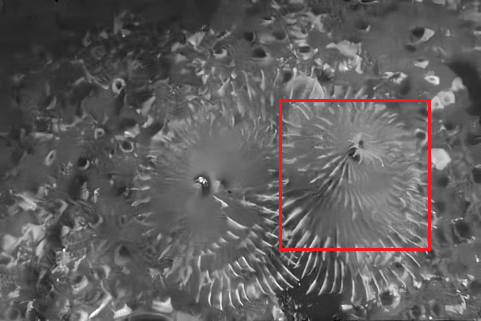}}\hfill
\subfigure[]
{\includegraphics[width=0.36\linewidth]{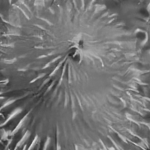}}\\
\caption{A denoising example to demonstrate the superiority of the
proposed nonlocal variant of TNRD over its local version
in the case of $\sigma=25$.
(a) TNRD (28.29dB, 0.7758), (b) cropped image.
(b) the proposed TNLRD model (28.39dB, 0.7821), (d) cropped image.
The numbers in the blankets are PSNR and SSIM values respectively.
One can see that the nonlocal variant produces less artifacts
in the smooth regions and more tiny details in the textured region, \eg
the zoomed-in part.}\label{fig:localVSnonlocal}
\end{figure}

Local methods concentrate on the modeling of the patch itself.
Nowadays, it is widely-known that another type of image prior is very
effective for image denoising - nonlocal self-similarity (NSS)
prior; that is, in natural images there are often many
similar patches (i.e., nonlocal neighbors) to a given patch,
which may be spatially far from it. Inspired by the seminal work of
nonlocal means \cite{buades2005non}, the NSS prior has been widely
exploited for image denoising in various framework, such as
K-SVD algorithm with nonlocal modeling \cite{mairaL2009non},
nuclear norm minimization with nonlocal modeling \cite{gu2014weighted}, and
Markov Random Fields with nonlocal modeling \cite{sun2011learning}.
Usually, NSS prior based models can significantly improve
their corresponding local versions. As a consequence,
many state-of-the-art image
denoising algorithms are built on the NSS prior, such as
BM3D \cite{dabov2007image}, LSSC \cite{mairaL2009non},
NCSR \cite{dong2011centralized}, and WNNM \cite{gu2014weighted}.

Usually, local methods cannot perform very well when the noise
level is high, because the correlations between neighboring pixels
have been corrupted by the severe noise. Therefore, it is generally
believed that local models are not expected to compete with
those nonlocal models, especially those state-of-the-art ones,
in terms of restoration quality. However, with the help of
techniques from machine learning, a few local models, such as
opt-MRF \cite{chen2014insights}, Cascade Shrinkage Fields (CSF)
\cite{schmidt2014shrinkage}, and recently proposed Trainable
Non-linear Reaction Diffusion (TNRD) \cite{chen2015learning},
succeed achieving state-of-the-art denoising performance via
appropriate modeling and supervised learning. It is noticeable that
the TNRD model has demonstrated strongly competitive, even better
performance against the best-reported nonlocal algorithm - WNNM,
meanwhile with much higher computational efficiency.

As mentioned earlier, incorporating the NSS prior has succeeded to boost
many image denoising algorithms. Therefore, we are highly motivated
to introduce the NSS prior to the best-performing diffusion framework -
TNRD to investigate whether it can also boost the TNRD model as usual.

\subsection{Our contributions}
The goal of this paper is to embed the NSS prior into the TNRD model
for the task of image denoising. To this end, we propose
trainable non-local reaction diffusion (TNLRD) models. The contributions
of this study are four-fold:
\begin{itemize}[leftmargin=*]
\setlength\itemsep{0em}
\item[a)]
We propose a compact matrix form to exploit the NSS prior, which
can facilitate the subsequent formulations and derivations
associated with the nonlocal modeling. In this work, the NSS prior is
defined by a set of non-local filters.
In a TNLRD model, the filter responses of $L$ similar patches
generated by a local spatial filter are further filtered by
its corresponding non-local filter.
\item[b)]
We construct the nonlocal diffusion process with fixed $T$ iterations,
which is parameterized by iteration-varying local spatial filters,
non-local filters and nonlinear influence functions.
Deriving the gradients of the training loss function
w.r.t those learning parameters is not trivial, due to the involved
nonlocal structure. We provide detailed derivations, which greatly
differ from the original TNRD model.
\item[c)]
The training phase is accomplished in a loss-specific manner,
where a loss function measuring the difference between clean
image $u_{gt}$ and denoised image $u_T$ is utilized to optimize the
model parameters. In this study, we investigate two different loss functions,
namely PSNR-oriented quadratic loss and SSIM related loss.
\item[d)]
We conduct comprehensive experiments
to demonstrate the denoising performance of the proposed TNLRD models.
As illustrated in Section \ref{sec:experiment},
the proposed TNLRD models outperform recent state-of-the-art methods
in terms of PSNR and SSIM.

\end{itemize}

The following section are organized as follows.
In Section 2, we give a brief review of the related works.
In Section 3, we introduce the proposed TNLRD models and the training issue.
In Section 4, we discuss the influence of the parameters in the proposed TNLRD models,
then show the denoising comparison with the previous state-of-the-arts.
Finally in Section 5, we draw the conclusion.
\section{Background and Related work}
In this section,
we first give a brief review of the TNRD model for image denoising,
then introduce the NSS scheme.
\subsection{Trainable non-linear reaction diffusion model}
Chen et. al \cite{chen2015learning} proposed
a simple but effective framework for image restoration - TNRD, which
is derived from the following energy functional
\begin{equation}\label{GibbsEnergy}
\text{E}(u|f) = \frac{\lambda}{2}\left| \left| u - f \right| \right|_2^2 + \overset{N_k}{\underset{i=1}\sum} \rho_i(k_i \ast u),
\end{equation}
where the regularization term
$\overset{N_k}{\underset{i=1}\sum} \rho_i(k_i \ast u)$
is a high-order MRFs - Fields of Experts (FoE) \cite{roth2009fields},
defined by a set of linear filters $k_i$ and the penalty function
$\rho_i$. $N_k$ is the number of filters, $*$ denotes the 2D convolution
operator. $\lambda$ is the strength of data term.

With appropriate modeling of the regularization term, minimizing
the energy functional \eqref{GibbsEnergy} can lead to a denoised
image. The steepest-descent procedure for minimizing
the energy \eqref{GibbsEnergy} reads as
\begin{equation}\label{steepest-descent}
\frac{u_t-u_{t-1}}{\Delta t} =
-\overset{N_k}{\underset{i=1}\sum} {\bar{k}_{i}} \ast \phi_{i}(k_{i} \ast u_{t-1}) - \lambda(u_{t-1} - f),
\end{equation}
where convolution kernel $\bar{k}_{i}$ is obtained by rotating the kernel $k_{i}$ 180 degrees,
$\phi_{i}(\cdot) = \rho_{i}{'}(\cdot)$ is
the influence function \cite{black1997robust} or flux function
\cite{weickert1998anisotropic}, $\Delta t$ donates the time step.

The TNRD model truncates the gradient descent procedure
\eqref{steepest-descent} to $T$ iterations, and it then naturally leads to
a multi-layer diffusion network with $T$ layers. This modification
introduces additional flexibility to the diffusion process, as
it becomes easier to train the influence function in this framework.
Moreover, as it can be considered as a multi-layer network,
we can exploit layer-varying parameters.
Therefore, the TNRD model is given as
the following diffusion network with $T$ layers.
\begin{equation}\label{TRDinferTstep}
\begin{cases}
u_0 = f,\quad t = 1, \cdots, T\\
u_t = u_{t-1} -
\left(
\overset{N_k}{\underset{i=1}\sum} {\bar{k}_{i}^{t}} \ast \phi_{i}^{t}(k_{i}^{t} \ast u_{t-1}) +
\lambda^t(u_{t-1} - f)
\right)\,.
\end{cases}
\end{equation}
Note that the parameters $\{k_{i}^{t}, \phi_{i}^{t}, \lambda^{t}\}$
vary across the layers. $f$ is the input of the diffusion network.
It is clear that each layer of
(\ref{TRDinferTstep}) only involves a few image convolution
operations, and therefore, it bears an
interesting link to the convolutional networks (CN)
employed for image restoration problems, such as \cite{jain2009natural}.

The parameters of TNRD models in (\ref{TRDinferTstep})
is trained in a supervised manner.
Given the pairs of noisy image $f$ and its ground-truth $u_{gt}$,
the parameters $\Theta^t =
\{k_{i}^{t}, \phi_{i}^{t}, \lambda^{t}\}$
are optimized by minimizing certain loss function
$\ell(u_T,u_{gt})$,
where $u_T$ is given by the inference
procedure (\ref{TRDinferTstep}). The training procedure is formulated as
\begin{equation}\label{TRDtrain}
\hspace{-0.2cm}\begin{cases}
\Theta^* = \text{argmin}_{\Theta}\cL(\Theta) = \sum\limits_{s = 1}^{S}\ell\left( u_T^s , u_{gt}^s \right) \\
\text{s.t.}
\begin{cases}
u_0^s = f^s \\
u_t^s = u_{t-1}^s -
\left(
\overset{N_k}{\underset{i=1}\sum} {\bar{k}_{i}^{t}} \ast \phi_{i}^{t}(k_{i}^{t} \ast u_{t-1}^s) +
\lambda^t(u_{t-1}^s - f^s)
\right)\\
t = 1 \cdots T\,,
\end{cases}
\end{cases}
\end{equation}
where $\Theta = \{\Theta^t\}_{t=0}^{t=T-1}$. The training problem can be solved via gradient
based algorithms, e.g., commonly used L-BFGS algorithm \cite{lbfgs}.
The gradients of the loss function with respect to $\Theta_t$ are computed
using the standard back-propagation technique widely used in
the neural networks learning \cite{lecun1998gradient}. There are two training strategies to learn the diffusion processes: 1) the greedy training strategy to learn the diffusion process stage-by-stage; and 2) the joint training strategy to joint train all the stages simultaneously. Generally speaking, the joint training strategy performs better \cite{chen2015learning}, and the greedy training strategy is often employed to provide a good initialization for the joint training. For simplicity, we just consider the joint training scheme to
train a diffusion process by simultaneously tuning
the parameters in all stages. The associated gradient $\frac {\partial \ell(u_T, u_{gt})}{\partial \Theta_t}$ is presented as follows,
\begin{equation}
\frac {\partial \ell(u_T, u_{gt})}{\partial \Theta_t} =
\frac {\partial u_t}{\partial \Theta_t} \cdot \frac {\partial u_{t+1}}{\partial u_{t}} \cdots
\frac {\partial \ell(u_T, u_{gt})}{\partial u_T} \,.
\label{iterstep}
\end{equation}

\subsection{Non-local self-similarity scheme}
\begin{figure}[t!]
\centering
\includegraphics[width=0.3\linewidth]{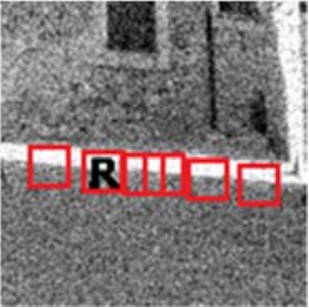}
\caption{Non-local similar patches of reference patch from noisy image corrupted by white Gaussian nise with $\sigma=25$.
The reference patch is marked with "R", non-local similar patches are also illustrated in red rectangles.}\label{fig:NSS}
\end{figure}

Based on the observation that one can always find a few similar patches
to a reference patch in the same image, which might be significantly
apart from the reference patch, an image prior named
non-local self-similarity (NSS) was introduced in
\cite{buades2005non}. As described in Fig. \ref{fig:NSS},
similar patches to a reference patch can be found in a significantly
larger spatial range than the patch size.
The non-local similar patches can be collected by
using a k-nearest neighbor (k-NN) algorithm \cite{sun2011learning}
or using a kernel function to map the patch distance to
coefficients \cite{dong2013sparse}.

The NSS prior has proven highly effective for many image
restoration problems, and it becomes greatly popular nowadays.
A lot of state-of-the-art image restoration algorithms
exploit this type of image prior, such as
image denoising algorithms BM3D \cite{dabov2007image} and WNNM \cite{gu2014weighted},
image interpolation approaches NARM \cite{dong2013sparse} and ANSM \cite{romano2014single}.
As a consequence, many local models also attempt to incorporate
the NSS prior to boost the performance of the local versions, such
as the LSSC method \cite{mairaL2009non}, which is a nonlocal
extension of the K-SVD algorithm.

We want to especially emphasize the a NSS prior induced method -
the NLR-MRF model proposed in \cite{sun2011learning},
which extends the spatial range of the original FoE model
\cite{roth2009fields}, as it is highly related to our work.
As described in \cite{sun2011learning}, in the NLR-MRF model,
several similar patches are firstly collected for each reference patch,
and then the responses of these similar patches
to a local filter are filtered by a cross-patch filter,
generating more sparse responses compared with the local filter
responses.
With the extend spatial range, NLR-MRF models surpassed
the original FoE models in both quality and quantity performance.

Our NSS prior extended TNRD model is also derived from a FoE prior based
model. Compared with the TNLRD model to exploit in this
paper, the NLR-MRF model is much more constrained in two aspects:
\begin{itemize}[leftmargin=*]
\setlength\itemsep{0em}
\item[a)] It employs unchanged parameters for each iteration. However,
our NLRD model makes use of iteration-varying parameters.
\item[b)] Although the penalty functions in the NLR-MRF model are
adjustable, they are functions of fixed shape (heavy-tailed functions
with a single minimum at the point zero), such as
Gaussian Scale Mixtures (GSM) or Student-t distribution.
In the TNLRD model, the influence functions are parameterized via radial
basis functions, which is able to generate functions of arbitrary shapes.
As demonstrated in \cite{chen2015learning}, those seemingly unconventional
influence functions found by the training phase play a key role
for the success of the TNRD model.
\end{itemize}

\section{Trainable non-local reaction diffusion models for image denoising}
In this section, we first describe the non-local filter,
then introduce the trainable non-local reaction diffusion for
image denoising, coined as TNLRD.
Finally we give the gradient derivation in the training issue.

\subsection{Compact matrix form to model the NSS prior}
In this work, we make use of k-NN algorithm to collect a
fixed number of similar patches.
Similar patches are collected by block matching
with mean squared error as patch similarity metrics
in a large searching window.
For the sake of computational efficiency, the size of searching window
is set to be several times larger than that of the local spatial filters,
as that in \cite{dabov2007image, sun2011learning, gu2014weighted}. For
each possible patch in an image $u$ of size $N \times M$
($p = N \times M$, and $u$ is represented as a column vector
$u \in \mathcal{R}^{p}$),
we collect $L$ similar patches (including the
reference patch itself) via block matching.
Therefore, after running block matching, we can
obtain results summarized in Table \ref{similar_patches}.
\begin{table}[t]
\vspace*{0.2cm}
\centering
\begin{tabular}{|c |c |c |c |c |c |c | c |}
\hline
index & 1 & 2 &3 &$\cdots$ & $n$ &$\cdots$ &$p$\\
\hline
$Q_2$ &$q_{12}$ & $q_{22}$ & $q_{32}$ &$\cdots$ & $q_{n2}$
&$\cdots$ &$q_{p2}$\\
\hline
$Q_3$ &$q_{13}$ & $q_{23}$ & $q_{33}$ &$\cdots$ & $q_{n3}$
&$\cdots$ &$q_{p3}$\\
\hline
$\cdots$ & $\cdots$& $\cdots$ & $\cdots$ &$\cdots$ & $\cdots$
&$\cdots$ &$\cdots$\\
\hline
$Q_j$ &$q_{1j}$ & $q_{2j}$ & $q_{3j}$ &$\cdots$ & $q_{nj}$
&$\cdots$ &$q_{pj}$\\
\hline
$\cdots$ & $\cdots$& $\cdots$ & $\cdots$ &$\cdots$ & $\cdots$
&$\cdots$ &$\cdots$\\
\hline
$Q_L$ &$q_{1L}$ & $q_{2L}$ & $q_{3L}$ &$\cdots$ & $q_{nL}$
&$\cdots$ &$q_{pL}$\\
\hline
\end{tabular}
\vspace*{0.2cm}
\caption{Block matching results}\vspace*{-0.5cm}
\label{similar_patches}
\end{table}

In Table \ref{similar_patches}, the numbers in each column
indicate the indexes of the found similar patches to the corresponding
reference patch. For example, in the column $n$, the numbers
$\{q_{n2}, q_{n3}, \cdots, q_{nL}\}$ indicate the indexes of $L-1$ similar
patches to the reference patch $n$, and the similar patches are
sorted according to the distance to the reference patch, \ie,
$d(P_{q_{n2}}, P_n) \leq d(P_{q_{n3}}, P_n) \leq \cdots \leq
d(P_{q_{nL}}, P_n)$, where $P_{q_{n2}}$ denotes an image patch centered
at the point $q_{n2}$, and function $d$ is a distance measurement of two
image patches.

Based on the results in Table \ref{similar_patches}, we construct $L$
highly sparse matrices of size $p \times p$, namely,
$\{V_1, V_2, \cdots, V_j, \cdots, V_L\} \in \mathcal{R}^{p \times p}$.
$V_j$ only involves the information from the $j^{th}$ row of
Table \ref{similar_patches}. Each row of $V_j$ contains merely a non-zero
number (exactly one), and its position is given by one of the
indexes $Q_j$. For example, in the $n^{th}$ row of matrix $V_j$,
only the element at position $q_{nj}$ is one, and the remaining elements
are all zeros. It is easy to see that matrix $V_1$ is the identity matrix,
\ie, $V_1 = I  \in \mathcal{R}^{p \times p}$.
As shown later, the NSS prior can be easily embedded into the TNRD
framework with the help of matrices $V_j$.

In our work,
we introduce a set of non-local filters
to embed NSS priors into the TNRD model.
A non-local filter is represented as a vector with $L$ elements,
\eg, $a = \{a_1, a_2, \cdots, a_L\}$,
whose $j^{th}$ value $a_j$ is assigned to the $j^{th}$ similar patch.
In the TNLRD model,
the filter response map $v_k$ generated by a spatial filter $k$
(\ie, $v_k = u * k$) is further filtered by a non-local filter $a$,
resulting a response map $v_{ka}$, then
for the reference patch $n$, its non-local filter response is given
as
\[
v_{ka}(n) = \overset{L}{\underset{j=1}\sum} a_j \cdot v_k(q_{nj}) \,.
\]

It turns out that the above formulation
can be given in a more compact way,
which reads as
\[
v_{ka} = W_a \cdot v_k \,,
\]
where the matrix $W_a$ is defined by $V_j$ and the non-local filter
$a$, given as
\begin{equation}\label{wa}
W_a = \overset{L}{\underset{j=1}\sum} a_j V_j \,.
\end{equation}

In the following subsections, we will see that formulating
the NSS prior in the way of \eqref{wa} can significantly
simplify the corresponding formulations, thus easier to
understand and to follow, when compared to the formulations
in \cite{sun2011learning}. In addition,
the non-local filter $W_a$ in matrix form is also highly sparse, as
each row of $W_a$ only has $L$ non-zero elements.
As a result, $W_a$ can be efficiently stored via sparse matrix.

\subsection{Trainable non-local reaction diffusion}
Following the formulation in the previous subsection to exploit the NSS prior,
it is easy embed the NSS prior into the TNRD framework, and then we arrive at
our proposed trainable non-local reaction diffusion.

In order to explain our proposed TNLRD model more clearly, we start from the
following energy functional, which incorporates the NSS prior in a natural way
\begin{equation}\label{GibbsEnergyNLMRF}
\text{E}(u|f) = \frac{\lambda}{2}\left| \left| u - f \right| \right|^2 +
\overset{N_k}{\underset{i=1}\sum} \rho_i(W_i K_i u)\,,
\end{equation}
where $u,f \in \mathcal{R}^p$ is the latent image and the noisy image respectively,
and $p$ is the number of pixels in image.
The local convolution kernel $k_i$ in (\ref{GibbsEnergy}) is represented
as its corresponding matrix form
$K_i \in \mathcal{R}^{p \times p}$ that is a highly sparse matrix,
such that
\[
k_i * u \Leftrightarrow K_i u \,.
\]
$W_i$ is highly sparse matrix defined as in \eqref{wa} to model the NSS prior,
which is related to a non-local filter.

We follow the basic idea of TNRD that unfolds the gradient descent process as
a multi-layer network model with layer-varying parameters, to derive the
proposed TNLRD model. It is easy to firstly check that the gradient of the
energy functional \eqref{GibbsEnergyNLMRF} with respect to $u$ is given as
\begin{equation}\label{gradientNL}
\frac{\partial \text{E}}{\partial u} =
\overset{N_k}{\underset{i=1}\sum} {\bar{K}_{i}} {W_{i}}^{\top} \phi_{i}(W_{i} K_{i}
u) + \lambda (u - f)\,,
\end{equation}
where function $\phi_i$ is given as $\phi_i = \rho_i'$, matrix $\bar{K}_{i}$
is the matrix form related to the linear kernel $\bar k_i$, which is
obtained by rotating kernel $k_i$ 180 degrees \footnote{
It should be noticed that the exact formulation for the first matrix in
\eqref{gradientNL} should be $K_i^\top$. We make use of $\bar{K}_{i}$ to simplify
the model complexity. More details can be found in \cite{chen2015learning}}.

Therefore, our proposed non-local diffusion model is given as
the following multi-layer network with layer-varying parameters
\begin{equation}\label{TNLRDinferTstep}
\begin{cases}
u_0 = f,\quad t = 1, \cdots, T\\
u_t = u_{t-1} -\left( \overset{N_k}{\underset{i=1}\sum} {\bar{K}_{i}^{t}} {W_{i}^{t}}^{\top} \phi_{i}^{t}(W_{i}^{t} K_{i}^{t} u_{t-1}) + \lambda^t(u_{t-1} - f) \right)\,.
\end{cases}
\end{equation}
Note that the parameters in layer $t$ include local filters $k_i$, non-local filters
$a_i$ (\ie, matrix $W_i$), nonlinear functions $\phi_i$ and the trade-off parameter
$\lambda$. The parameter set in layer $t$ is given as
$\Theta_t = \{ \lambda^t,\phi_{i}^{t},K_{i}^{t},W_{i}^{t} \}$,
where $i=1,...,N_k$ and $t=1,...,T$.

According to the diffusion process \eqref{TNLRDinferTstep}, one can see
that the TNRD model can be treated as a special case of the TNLRD models with $L=1$,
as the corresponding non-local diffusion model clearly degenerates to the
local version given in \eqref{TRDinferTstep},
if we set $W = a_1 V_1 = I  \in \mathcal{R}^{p \times p}$.

In this work, we parameterize the local filters, non-local filters,
nonlinear functions in the following way. Concerning the local filters, we follow
the TNRD model, and exploit zero-mean filters of unit norm. This is accomplished by
constructing the filter $k_i$ in the way of
\begin{equation}\label{k_c}
k_i = \mathcal{B} \frac{c_i}{\|c_i\|_2},
\end{equation}
where $\|\cdot\|_2$ denotes the $\ell_2$-norm, and $\mathcal{B}$ is a filter bank.
Therefore, it is clear that the filter $k$ is a linear combination of the basis
filters in the filter bank. In order to achieve the property of zero-mean, the filter
bank $\mathcal{B}$ in this work is chosen as a modified DCT basis,
which is obtained by removing the filter with constant entries from the complete
DCT filters.

The non-local filters in the TNLRD models are vectors with unit length constraint.
Therefore, we construct the non-local filter $a_i$ as
\begin{equation}\label{a_b}
a_i = \frac{b_i}{\| b_i \|_2},
\end{equation}
where $b_i$ is completely free of any constraint.

Following the work of TNRD, the nonlinear functions
$\phi_{i}$ are parameterized via radial basis function (RBFs), i.e.,
function $\phi_i$ is represented as a weighted linear combination
of a set of RBFs as follows,
\begin{equation}\label{rbf}
\phi_{i}(z) = \overset{M}{\underset{j=1}\sum} \alpha_{ij} \varphi (\frac {\left| z - \mu_j \right|}{\gamma}),
\end{equation}
where $\varphi(\cdot)$ here is Gaussian RBFs
with equidistant centers $\mu_j$ and unified scaling $\gamma$. The
Gaussian radial basis is defined as
\[
\varphi \left(\frac {|z - \mu|}{\gamma}\right)  =
\text{exp}\left(-\frac {(z - \mu)^2}{2\gamma^2}\right)
\]

As described above, the proposed TNLRD model contains plenty of free parameters,
which can be learned from training samples. In this work,
we train the TNLRD model parameters in a loss-based learning manner.
Given the pairs of degraded image $f$ and the ground-truth original image $u_{gt}$,
the parameters are optimized by minimizing certain loss function
$\ell(u_T,u_{gt})$, which is defined to measure the difference between
the output $u_T$, given by the inference procedure (\ref{TNLRDinferTstep})
and the desired output, \ie, the ground-truth $u_{gt}$.

In summary, the training procedure is formulated as
\begin{equation}\label{NLMRFtrain}
\small
\hspace{-0.2cm}
\begin{cases}
\Theta^* = \text{argmin}_{\Theta}\cL(\Theta) = \sum\limits_{s = 1}^{S}\ell\left( u_T^s , u_{gt}^s \right) \\
\text{s.t.}
\begin{cases}
u_0^s = f^s \\
u_t^s = u_{t-1}^s - \left( \overset{N_k}{\underset{i=1}\sum}
{\bar{K}_{i}^{t}} {W_{i}^{t}}^{\top} \phi_{i}^{t}(W_{i}^{t} K_{i}^{t} u_{t-1}^s) +
\lambda^t(u_{t-1}^s - f^s) \right)\\
t = 1 \cdots T.
\end{cases}
\end{cases}
\end{equation}
where the parameters $\Theta = \{\Theta_t\}_{t=1}^T$. Note that we do not
specify the form of the loss function in the training phase at present.
The basic requirement for the loss function is that it should be
differentiable. In our study,
we consider two different loss functions for training, see Section
\ref{loss_function}.

\subsection{Gradients in the training phase for the TNLRD model}
Usually, gradient-based algorithms are exploited to solve the corresponding
optimization problem \eqref{NLMRFtrain} in the training phase.
Therefore, it is important to compute the gradients of cost function with respect to
model parameters $\Theta$ for TNLRD.

The gradient of loss function $\ell(u_T,u_{gt})$ with respect to
parameters in the layer $t$, \ie, $\Theta_t$,
is computed using back-propagation technique
widely used in neural networks learning \cite{lecun1998gradient},
\begin{equation}\label{NLMRFtrainParameter}
\frac{\partial \ell(u_T,u_{gt})}{\partial \Theta_t} = \frac{\partial u_t}{\partial \Theta_t} \cdot \frac{\partial u_{t+1}}{\partial u_t} \cdot \cdot \cdot \frac{\partial \ell(u_T,u_{gt})}{\partial u_T}.
\end{equation}

In the case of quadratic loss function, \ie,
\begin{equation}\label{NLMRFtrainLoss}
\ell(u_T,u_{gt}) = \frac{1}{2} \left| \left| u_T - u_{gt} \right| \right|_2^2,
\end{equation}

$\frac{\partial \ell(u_T,u_{gt})}{\partial u_T}$
is directly derived from (\ref{NLMRFtrainLoss}),
\begin{equation}\label{grad_l_u}
\frac{\partial \ell(u_T,u_{gt})}{\partial u_T} = u_T-u_{gt}.
\end{equation}

$\frac{\partial u_{t+1}}{\partial u_t}$ is computed from (\ref{TNLRDinferTstep}),
\begin{equation}\label{grad_u_u}
\frac{\partial u_{t+1}}{\partial u_t} = (1-\lambda^{t+1})I - \overset{N_k}{\underset{i=1}\sum} {K_{i}^{t+1}}^{\top} {W_{i}^{t+1}}^{\top} \Lambda_{i} W_{i}^{t+1} (\bar{K}_{i}^{t+1})^{\top},
\end{equation}
where matrix $\Lambda_{i}$ is a diagonal matrix given as $\Lambda_{i} = \text{diag}
({\phi_{i}^{t+1}}'(z_1),\cdots,{\phi_{i}^{t+1}}'(z_p))$,
$z=W_{i}^{t+1} K_{i}^{t+1} u_t$.

Combining the result of \eqref{grad_u_u} and \eqref{NLMRFtrainParameter}, we arrive
at the result of
$\frac{\partial \ell(u_T,u_{gt})}{\partial u_t}$, denoted by
\[
\frac{\partial \ell(u_T,u_{gt})}{\partial u_t} = e\,.
\]

Now, we focus on the computation of the gradients
$\frac{\partial u_t}{\partial \Theta_t}$, which are derived from
the diffusion procedure \eqref{TNLRDinferTstep}.

\subsubsection{Computing $\frac{\partial \ell}{\partial \lambda^t}$}

$\frac{\partial u_t}{\partial \lambda^t}$ is computed as
\begin{equation}\label{grad_u_lambda}
\frac{\partial u_t}{\partial \lambda^t} = -(u_{t-1}-f)^{\top}.
\end{equation}
Therefore, $\frac{\partial \ell}{\partial \lambda_t}$ is given as
\begin{equation}
\frac{\partial \ell}{\partial \lambda^t} = -(u_{t-1}-f)^{\top}e\,.
\end{equation}

\subsubsection{Computing $\frac{\partial \ell}{\partial c_i^t}$}
Firstly, $\frac{\partial u_t}{\partial k_i^t}$ is computed as
\begin{equation}\label{grad_u_k}
\frac{\partial u_t}{\partial k_i^t} = - \left( \Xi_{inv}^{\top} {Y}^{\top} + U_{t-1}^{\top} {W_{i}^{t}}^{\top} \Lambda_{i} W_{i}^{t} (\bar{K}_{i}^{t})^{\top} \right)\,,
\end{equation}
where matrix $\Lambda_{i}$ is a diagonal matrix given as $\Lambda_{i} = \text{diag}
({\phi_{i}^{t}}'(z_1),\cdots,{\phi_{i}^{t}}'(z_p))$,
$z=W_{i}^{t} K_{i}^{t} u_{t-1}$. Matrix $U_{t-1}$ and $Y$ are constructed from
the images $u_{t-1}$ and $y$ of 2D form, respectively.
For example, $U_{t-1}$ is constructed in the way that its rows are vectorized local
patch extracted from image $u_{t-1}$ for each pixel, such that
\[
k_i * u_{t-1} \Leftrightarrow  U_{t-1} k_i\,.
\]
The matrix $Y$ is defined in the same way, and the image $y$ is given as
$y={W_{i}^{t}}^{\top} \phi_{i}^{t}(W_{i}^{t} K_{i}^{t} u_{t-1})$.
Matrix $\Xi_{inv}^\top$ is a linear operator which inverts
the vectorized kernel $k$.
In the case of a square kernel $k$, it is equivalent to the Matlab command
\[
\Xi_{inv}^\top k \Longleftrightarrow rot90(rot90(k)) \,.
\]

As a consequence, $\frac{\partial \ell}{\partial k_i^t}$ is given as
\begin{equation}\label{grad_l_k}
\frac{\partial \ell}{\partial k_i^t} =
- \left( \Xi_{inv}^{\top} {Y}^{\top} + U_{t-1}^{\top} {W_{i}^{t}}^{\top} \Lambda_{i} W_{i}^{t} (\bar{K}_{i}^{t})^{\top} \right) e\,.
\end{equation}

As the filter $k_i$ is parameterized by coefficients of $c_i$, we need to additionally
calculate $\frac{\partial k_i^t}{\partial c_i^t}$, which is
computed from (\ref{k_c}),
\begin{equation}\label{grad_k_c}
\frac{\partial k_i^t}{\partial c_i^t} = \frac{1}{{\left| \left| c_i^t \right| \right|}_2} \left(I - \frac{c_i^t}{{\left| \left| c_i^t \right| \right|}_2} \cdot \frac{(c_i^t)^{\top}}{{\left| \left| c_i^t \right| \right|}_2}\right) \cdot \mathcal{B}^{\top}.
\end{equation}
Combining the results of \eqref{grad_k_c} and \eqref{grad_l_k}, we can obtain
the required gradients $\frac{\partial \ell}{\partial c_i^t}$.

\subsubsection{Computing $\frac{\partial \ell}{\partial \alpha_i^t}$}
$\frac{\partial u_t}{\partial \alpha_i^t}$ is computed from
(\ref{TNLRDinferTstep}) and (\ref{rbf}),
\begin{equation}\label{grad_u_varphi}
\frac{\partial u_t}{\partial \alpha_i^t} = -G^{\top} W_i^t (\bar{K}_{i}^{t})^{\top},
\end{equation}
where $G_{rc}=\varphi (\frac {\left| z_r - \mu_c \right|}{\gamma})$,
$G \in \mathcal{R}^{p \times M}$, $r = 1,\cdots, p$ and $c = 1,\cdots, M$,
$z=W_{i}^{t} K_{i}^{t} u_{t-1}$.

Therefore, $\frac{\partial \ell}{\partial \alpha_i^t}$ is given as
\begin{equation}
\frac{\partial \ell}{\partial \alpha_i^t} =
-G^{\top} W_i^t (\bar{K}_{i}^{t})^{\top} e\,.
\end{equation}

{
\subsubsection{Computing $\frac{\partial \ell}{\partial b_i^t}$}

Firstly, $w_i^t$ and $\bar{w}_i^t$ are defined as the vectorized form of matrix $W_i^t$ and ${W_i^t}^{\top}$ respectively, holding that
$w_i^t = \mathbf{vec}(W_i^t) \in \mathcal{R}^{p^2}$ and
$\bar{w}_i^t = \mathbf{vec}({W_i^t}^{\top}) \in \mathcal{R}^{p^2}$.
The relation between $w_i^t$ and $\bar{w}_i^t$ reads as
\[
\bar{w}_i^t = P w_i^t \,,
\]
where matrix $P$ is a rearrange matrix. 

$\frac{\partial u_t}{\partial w_i^t}$ is computed from (\ref{TNLRDinferTstep}), 
\[
\frac{\partial u_t}{\partial w_i^t} = - \frac{\partial \bar{K}_{i}^{t} {W_{i}^{t}}^{\top} \phi_{i}^{t}(W_{i}^{t} K_{i}^{t} u_{t-1})}{\partial w_i^t}
 = - \frac{\partial y}{\partial w_i^t} (\bar{K}_{i}^{t})^{\top} ,
\]
where $y={W_{i}^{t}}^{\top} h$, $h = \phi_{i}^{t}(z)$, $z = W_{i}^{t} \hat{u}$, and $\hat{u} = K_{i}^{t} u_{t-1}$.
In the computation of $\frac{\partial y}{\partial w_i^t}$, the following relations are useful, namely
\[
y = {W_{i}^{t}}^{\top} h = H \bar{w}_i^t
\]
and
\[
z = W_{i}^{t} \hat{u} = \hat{U} w_i^t,
\]
where matrix $H$ and $\hat{U}_{t-1}$ are highly sparse, given as
\[
H =
\begin{pmatrix}
h_1 \cdots 0 \, h_2 \cdots 0 \cdots h_p \cdots 0\\
0 \, h_1 \cdots 0 \, h_2 \cdots \cdots 0 \, h_p \cdots \\
\vdots \\
0 \cdots h_1 \, 0 \cdots h_2 \cdots 0 \cdots h_p
\end{pmatrix}
\]
and
\[
\hat{U} =
\begin{pmatrix}
\hat{u}_1 \cdots 0 \, \hat{u}_2 \cdots 0 \cdots \hat{u}_p \cdots 0\\
0 \, \hat{u}_1 \cdots 0 \, \hat{u}_2 \cdots \cdots 0 \, \hat{u}_p \cdots \\
\vdots \\
0 \cdots \hat{u}_1 \, 0 \cdots \hat{u}_2 \cdots 0 \cdots \hat{u}_p
\end{pmatrix}
\]
respectively. 

Given that $y = {W_{i}^{t}}^{\top} h = H \bar{w}_i^t$,
$\frac{\partial y}{\partial w_i^t}$ is computed as
\[
\frac{\partial y}{\partial w_i^t} = \frac{\partial h}{\partial w_i^t} \cdot \frac{\partial y}{\partial h} + \frac{\partial \bar{w}_i^t}{\partial w_i^t} \cdot \frac{\partial y}{\partial \bar{w}_i^t} \\
= \frac{\partial h}{\partial w_i^t} W_i^t + P^{\top} H^{\top},
\]
where $\frac{\partial h}{\partial w_i^t}$ is given as
\[
\frac{\partial h}{\partial w_i^t} = \frac{\partial z}{\partial w_i^t} \Lambda_{i} = {\hat{U}}^{\top} \Lambda_{i},
\]
where $\Lambda_{i} = \text{diag}
({\phi_{i}^{t}}'(z_1),\cdots,{\phi_{i}^{t}}'(z_p))$.

Combining these derivation, $\frac{\partial u_t}{\partial w_i^t}$ is computed as
\begin{equation}\label{grad_u_w}
\frac{\partial u_t}{\partial w_i^t} = -\left( P^{\top} H^{\top} + {\hat{U}}^{\top} \Lambda_{i} W_{i}^{t} \right) (\bar{K}_{i}^{t})^{\top}.
\end{equation}

As $W_i^t$ is computed from (\ref{wa}), $\frac{\partial w_i^t}{\partial a_{i}^t}$ is given as,
\begin{equation}\label{grad_w_a}
\frac{\partial w_i^t}{\partial a_{i}^t} =
\begin{pmatrix}
v_1^{\top}\\
v_2^{\top}\\
\vdots\\
v_L^{\top}
\end{pmatrix} \,,
\end{equation}
where $v_j = \mathbf{vec}(V_j) \in \mathcal{R}^{p^2}, j=1, 2, \cdots, L$.
Then, we can obtain $\frac{\partial \ell}{\partial a_i^t}$ from (\ref{grad_u_w}) and (\ref{grad_w_a}), given as
\begin{equation}\label{grad_l_a}
\frac{\partial \ell}{\partial a_i^t} =
\frac{\partial w_i^t}{\partial a_i^t} \cdot \frac{\partial u_t}{\partial w_i^t} \cdot e\,.
\end{equation}

As the non-local filter $a_i$ is parameterized by the coefficients
$b_i$ as shown in (\ref{a_b}), we need to additionally compute
$\frac{\partial a_{i}^t}{\partial b_{i}^t}$
\begin{equation}\label{grad_a_b}
\frac{\partial a_{i}^t}{\partial b_{i}^t} = \frac{1}{{\left| \left|  b_{i}^t \right| \right|}_2} \left(I - \frac{b_{i}^t}{{\left| \left|  b_{i}^t \right| \right|}_2} \cdot \frac{(b_{i}^t)^{\top}}{{\left| \left| b_{i}^t \right| \right|}_2}\right).
\end{equation}

Combining the gradients in \eqref{grad_a_b} and \eqref{grad_l_a}, we can obtain
the required gradient $\frac{\partial \ell}{\partial b_i^t}$.

The direct computation of $\frac{\partial \ell}{\partial a_i^t}$ is quite time-consuming and memory-inefficient.
Benefit from the sparse matrix structure,
matrix $H$, $\hat{U}$ and $\frac{\partial w_i^t}{\partial a_i^t}$ are not constructed explicitly.
Therefore, the computation of $\frac{\partial \ell}{\partial a_i^t}$
is quite efficient.
As mentioned above, matrix $\{V_j\}_{j=1}^T$ are highly sparse,
each row of $V_j$ has precisely one non-zero value, and others are zeros.
Therefore, the computation of $\frac{\partial \ell}{\partial a_i^t}$ can be interpreted as picking up values from $\frac{\partial \ell}{\partial w_i^t}$ indexed by $\frac{\partial w_i^t}{\partial a_i^t}$.
The computation of $\frac{\partial \ell}{\partial w_i^t}$ can be further simplified as
\[
\frac{\partial \ell}{\partial w_i^t} = - h {\hat{e}}^{\top} - (diag(W_i^t \hat{e}) h') {\hat{u}}^{\top},
\]
where $\hat{e} = (\bar{K}_{i}^{t})^{\top} e \in \mathcal{R}^{p}$, $h'$ is the derivative of $h$ w.r.t. $z$.
Considering the sparse structure of $V_j$, the indexing of $\frac{\partial \ell}{\partial w_i^t}$ is actually the indexing of $\hat{u}$ and $\hat{e}$ using $\frac{\partial w_i^t}{\partial a_i^t}$ in forms that described in Table \ref{similar_patches}.
The computational complexity of $\frac{\partial \ell}{\partial a_i^t}$ can be greatly reduced.

Implementation will be made publicly available after acceptance.
}

\section{Experimental results}\label{sec:experiment}
\subsection{Training of TNLRD models}
Concerning the model complexity
\footnote{While TNLRD models with more stages provide better denoising performance,
they cost more time in both training and inference phase.},
the stages of inference $T$ is set to 5;
the local filter size $m \times m$ is set to $5 \times 5$ and $7 \times 7$;
the number of non-local similar patches $L$ is set to 3, 5, 7, 9.
The size of searching window is $31 \times 31$.
The size of block matching is $7 \times 7$.

We trained the TNLRD models for Gaussian denoising problem with different standard deviation $\sigma$.
We minimize (\ref{NLMRFtrain}) to learn the parameters of the TNLRD models
with commonly used gradient-based L-BFGS \cite{liu1989limited}.
The gradient of loss function with respect to parameters can be derived from (\ref{NLMRFtrainParameter}) - (\ref{grad_a_b}).
The training dataset of original and noisy image pairs
is constructed over 400 images as \cite{chen2015learning} \cite{schmidt2014shrinkage}.
We cropped a $180 \times 180$ region from each image,
resulting in a total of 400 training images of size $180 \times 180$.
In the training phase,
computing the gradients of one stage for 400 images of size $180 \times 180$ takes about 480$s$
on a server with CPUs: Intel(R) Xeon E5-2650 @ 2.00GHz (eight cores).
We run 200 L-BFGS iterations for optimization.
Therefore, the total training time for
${\text{TNLRD}}_{7 \times 7 \times 3}^{5}$ model is $5 \times (200 \times 480)/3600 = 133.3h$.
Implementation will be made publicly available after acceptance.

\begin{figure*}[t!]
\centering
\includegraphics[width=1.0\linewidth]{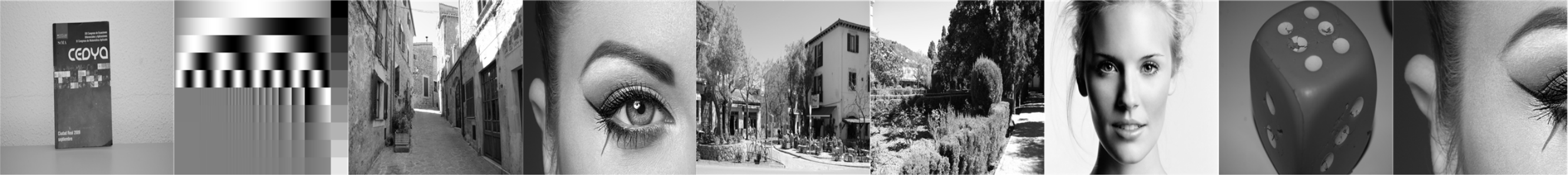}
\caption{9 image set which are collecting from web.}\label{fig:9imgset}
\end{figure*}

In order to perform a fair comparison to previous works, i.e.,
BM3D \cite{dabov2007image},
WNNM \cite{gu2014weighted},
NLR-MRF \cite{sun2011learning}
and TNRD \cite{chen2015learning},
we used the 68 test images in \cite{chen2015learning},
which are original introduced by \cite{roth2009fields}
and are widely used in image denoising.
We evaluated the denoising performance using
PSNR as \cite{chen2015learning} and SSIM as \cite{Wang04imagequality}.
SSIM provides a perceptually more plausible image error measure,
which has been verified in psychophysical experiments.
SSIM values range between 0 and 1, where 1 is a perfect restoration.
We also test TNRD and our TNLRD models on a 9 image set which are collecting from web,
as shown in Fig. \ref{fig:9imgset}.
The codes of the comparison methods were downloaded from the authors's homepage.
					
\subsection{Influence of parameters initialization}
The TNLRD models with different parameters configuration are denoted as,
${\text{TNLRD-}method}_{m \times m \times L}^{T}$.
The $method$ denotes the parameters initialization method,
$tnrd$ for initializing from the TNRD models,
and $plain$ for initializing from plain settings.
In \cite{sun2011learning},
the author trained NLR-MRF models starting from
MRF models with local spatial clique, i.e., FoE models, using NSS setting.
We followed the same training scheme as that in \cite{sun2011learning} for
training NLR-MRF. We started from the local TNRD models by setting $a_{i1}^t=1$
and $a_{ij}^t=0$, $j=2,...,L$,
and conducted a joint training for parameters of the $T$ steps inference (\ref{TNLRDinferTstep}),
denoted as ${\text{TNLRD-}tnrd}_{m \times m \times L}^{T}$.

We also trained the parameters of TNLRD models via the greedy
training from plain initialization,
then jointly trained the $T$ steps inference (\ref{TNLRDinferTstep}),
denoted as ${\text{TNLRD-}plain}_{m \times m \times L}^{T}$.
Greedy training means a strategy that greedily trains a multi-layer diffusion network
layer by layer. In the plain initialization training,
we observed that TNLRD models with joint training surpass
models obtained in greedy training by 0.55dB in average. Therefore,
it is recommended that joint training should be conducted after greedy training.

We trained TNLRD models using both parameters initialization method,
and got two models,
namely ${\text{TNLRD-}tnrd}_{7 \times 7 \times 5}^{5}$
and ${\text{TNLRD-}plain}_{7 \times 7 \times 5}^{5}$.
We evaluated their denoising performance on the 68 test images.
Models trained by $tnrd$ and $plain$ initialization
achieve almost the same denoising performance, i.e., 29.01dB in average.
This conclusion holds for our models with other model capacities.
For the sake of training efficiency
\footnote{The plain initialization with greedy and joint training is
more time consuming than the TNRD initialization which only conducts joint training.},
in the following experiments,
we mainly discuss the models trained via TNRD initialization,
which is coined as ${\text{TNLRD}}_{m \times m \times L}^{T}$
omitting the $method$ in ${\text{TNLRD-}method}_{m \times m \times L}^{T}$.

\subsection{Influence of number of non-local similar patches}
In this subsection, we investigate the influence of
different number of non-local similar patches $L$ for
both ${\text{TNLRD}}_{5 \times 5 \times L}^{5}$
and ${\text{TNLRD}}_{7 \times 7 \times L}^{5}$.

\begin{figure}[t!]
\centering
\includegraphics[width=0.9\linewidth]{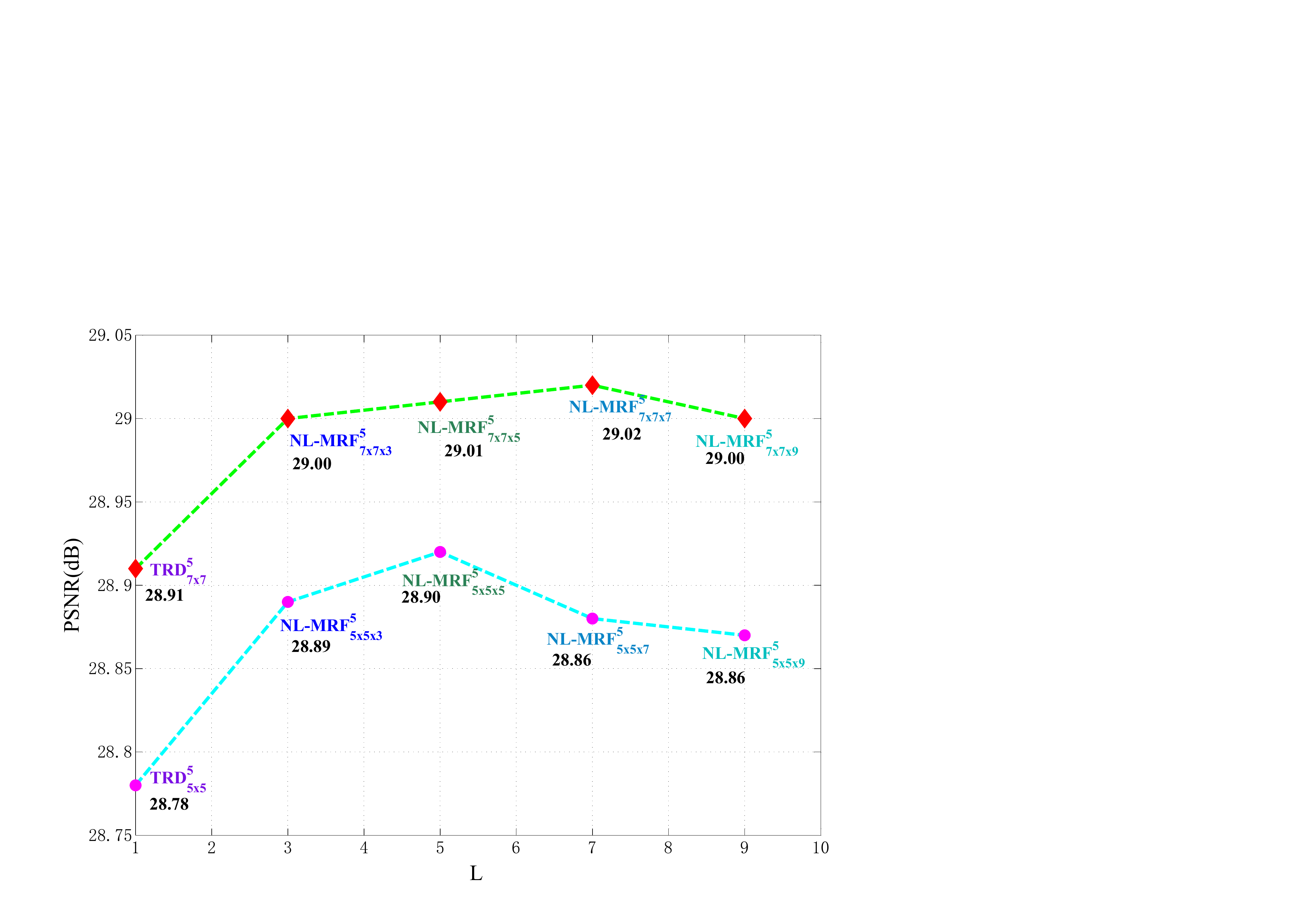}
\caption{Influence of the number of non-local similar patches $L$ and filter size $m \times m$.
The green dash line with red diamond markers donates the performance of filter size with $7 \times 7$,
and cyan dash line with magenta circle markers denates the performance of filter size with $5 \times 5$.
TNRD can be regarded as a specical case of TNLRD with $L=1$.}\label{fig:NLNumber}
\end{figure}

As described above, the TNRD model can be treated as a special case of
the TNLRD model with $L=1$. Therefore, in the training phase, the TNLRD model can
be initialized from its local version. The denoising performance of
the trained TNLRD models with different configurations are illustrated in
Fig. \ref{fig:NLNumber}.
One can see that the performance of the trained models is improved when $L$ increases,
and is degraded when $L$ continues to increase.
A performance peak exists.
for ${\text{TNLRD}}_{5 \times 5 \times L}^{5}$, it is $L=5$;
for ${\text{TNLRD}}_{7 \times 7 \times L}^{5}$, it is $L=7$.
While a peak exists, the performance gap is within 0.05dB.
For the sake of computational efficiency
\footnote{Larger $L$ will take more time for both training phase and
test phase.},
in the rest of this section, we set $L=5$.
${\text{TNLRD}}_{5 \times 5 \times 5}^{5}$
surpasses ${\text{TNRD}}_{5 \times 5}^5$ about 0.14dB.
${\text{TNLRD}}_{7 \times 7 \times 5}^{5}$
surpasses ${\text{TNRD}}_{7 \times 7}^5$ about 0.10dB.

\begin{figure}[t!]
\centering
\subfigure[48 local filters of size $7 \times 7$ and the corresponding non-local filters of size $1 \times 5$ in stage 1]
{\includegraphics[width=1\linewidth]{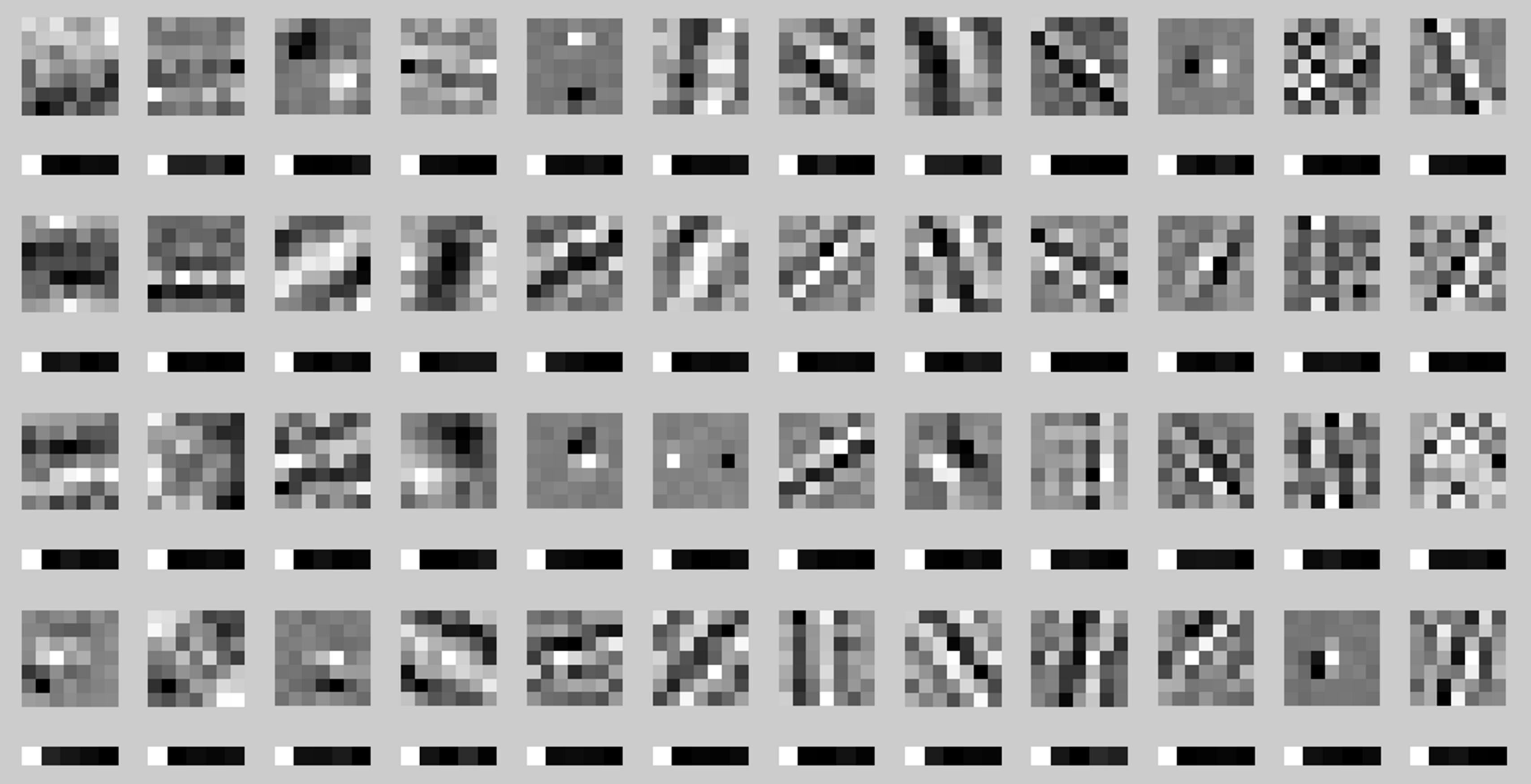}}
\subfigure[48 local filters of size $7 \times 7$ and the corresponding non-local filters of size $1 \times 5$ in stage 5]
{\includegraphics[width=1\linewidth]{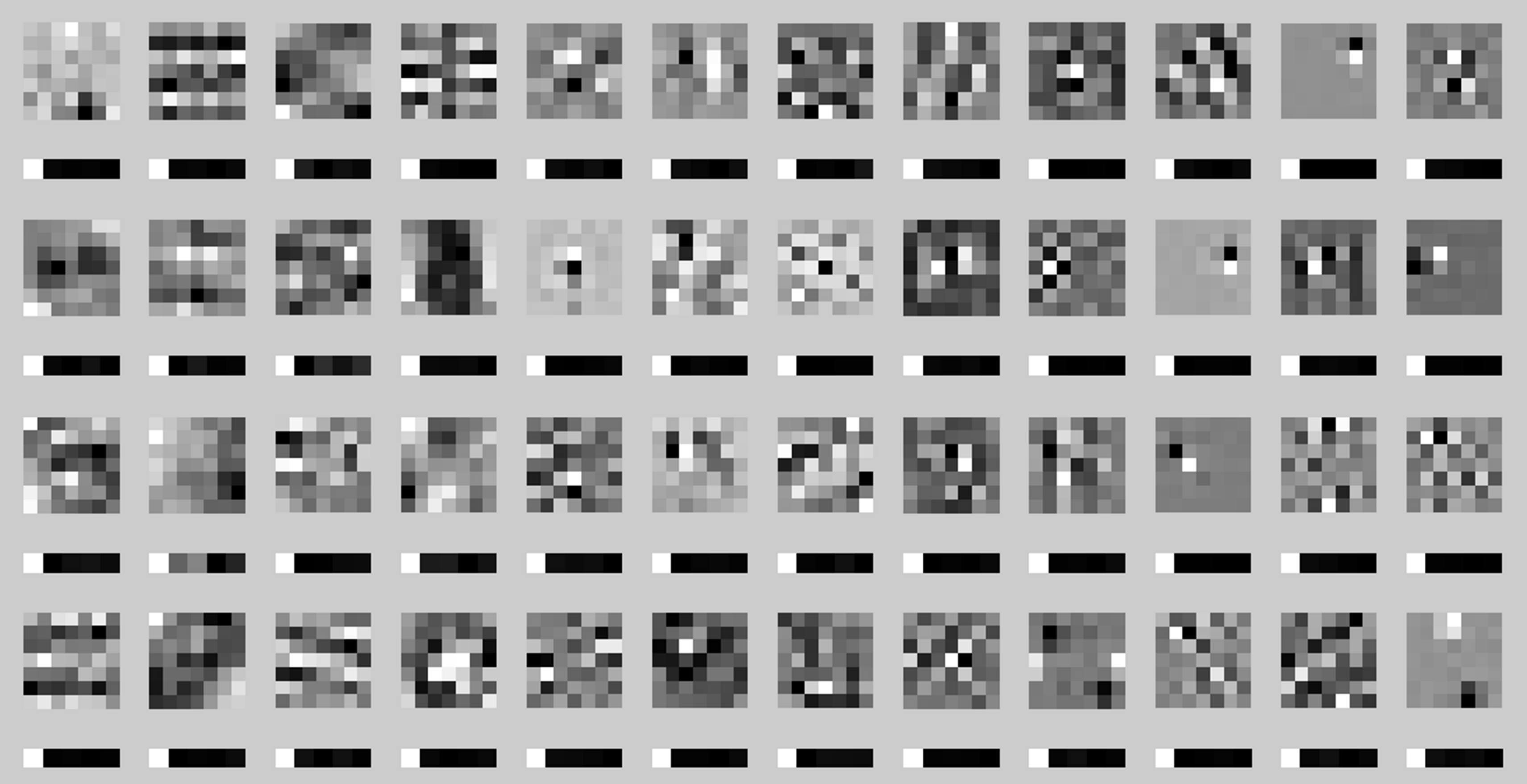}}
\caption{Trained local and non-local filters (in the first and last stage) of
${\text{TNLRD}}_{7 \times 7 \times 5}^{5}$ model for the noise level $\sigma = 25$.
The 1-D vectors below the local filters are the corresponding non-local filters.
In most of the non-local filters, the first element is near 1, while rest are near zero.}\label{fig:trainedFilters}
\end{figure}

Fig. \ref{fig:trainedFilters} shows the trained local
and non-local filters of
${\text{TNLRD}}_{7 \times 7 \times 5}^{5}$
in the first and last inference stage,
in the training of Gaussian denoising with $\sigma = 25$.
In most of the non-local filters,
the first element is near 1, while the rest are near zero,
for example $[0.999, 0.001, -0.018, -0.010, -0.027]^{\top}$;
while in some of the non-local filters,
the first element is of the same scale with the rest,
for example $[0.915, 0.185, 0.194, 0.178, 0.243]^{\top}$.
The former non-local filters are related with simple local filters,
for example the directional derivatives.
The later non-local filters are related with complex local filters,
hence all the local filter response of the similar patches are useful.

\subsection{Influence of filter size}
We also investigate the influence of filter size, as shown in Fig. \ref{fig:NLNumber}.
The increasing of the filter size from $5 \times 5$ to $7 \times 7$
brings an average 0.11dB improvement.
In the evaluating of denoising performance,
we prefer ${\text{TNLRD}}_{7 \times 7 \times 5}^{5}$ model
as it provides better trade-off between performance and run time.

\subsection{Influence of loss function}\label{loss_function}
\begin{table}[t!]
\caption{Influence of different loss function L2 and SSIM for image denoising with $\sigma=15$, $25$ and $50$.
Average PSNR (dB) and SSIM on 68 images for each $\sigma$.}\label{table:lossfunc}
\centering
\begin{tabular}{|c|cc|cc|cc|}
\hline
\multirow{3}{*}{Inference} & \multicolumn{6}{c|}{Training} \\
\cline{2-7}
{} & \multicolumn{2}{c|}{$\sigma=15$} & \multicolumn{2}{c|}{$\sigma=25$} & \multicolumn{2}{c|}{$\sigma=50$}\\
\cline{2-7}
{} & L2 & SSIM & L2 & SSIM & L2 &  SSIM \\
\hline
PSNR & \textbf{31.50} & 0.8852 & \textbf{29.01} & 0.8201 & \textbf{26.06} & 0.7094 \\
SSIM & 31.31 & \textbf{0.8864} & 28.83 & \textbf{0.8219} & 25.80 & \textbf{0.7113} \\
\hline
\end{tabular}
\end{table}

In \cite{yu2015inpainting, Zhao2015L2},
the loss function for discriminative training is SSIM
instead of L2 for image inpainting and denoising respectively.
The trained models with SSIM loss function may provide visually more plausible results.
Inspired by these works, we trained our TNLRD models using SSIM loss function as \cite{Wang04imagequality}.
In the case of $\sigma = 25$, the trained TNLRD models via
SSIM loss achieves SSIM result of 0.8219,
while the corresponding average PSNR is 28.83dB,
as shown in Table \ref{table:lossfunc}.
The TNLRD models with the same capacity trained via the L2 loss,
achieves a result of SSIM = 0.8201 and PSNR = 29.01dB.
As shown in Fig. \ref{fig:68img12},
the TNLRD models trained via SSIM loss offer sharper image than that trained via L2 loss.
SSIM loss function benefits the TNLRD models to produce more visually plausible denoising
results. From Table \ref{table:68imagesPSNR},
we note that our TNLRD models trained via SSIM loss
achieve competitive performance with WNNM in terms of PSNR,
and provide better recovered images in terms of SSIM.
We also note that, compared with models trained via SSIM loss,
models trained via L2 loss achieve competitive performance in terms of SSIM,
and superior performance in terms of PSNR.
Bearing these in mind,
in the following comparison with other image denoising methods,
we prefer the models trained with L2 loss.

\subsection{Denoising}
\begin{figure*}[t!]
\centering
\subfigure[original]
{\includegraphics[width=0.24\linewidth]{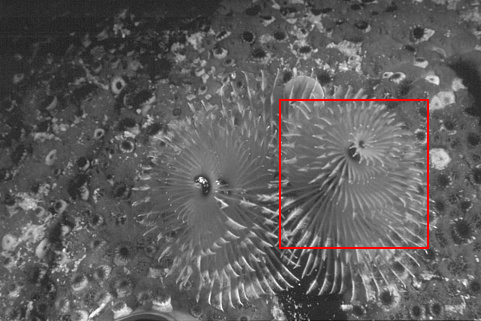}}\hfill
\subfigure[noisy]
{\includegraphics[width=0.24\linewidth]{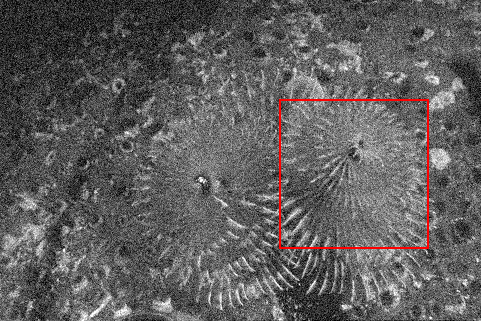}}\hfill
\subfigure[BM3D (27.98dB, 0.7642)]
{\includegraphics[width=0.24\linewidth]{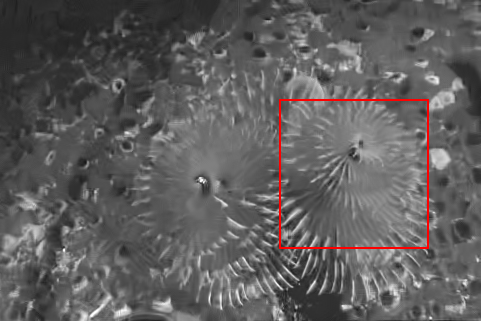}}\hfill
\subfigure[WNNM (28.24dB, 0.7706)]
{\includegraphics[width=0.24\linewidth]{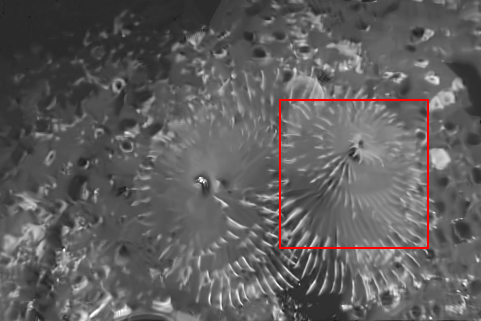}}\\
\vspace{2mm}
\subfigure[NLR-MRF (27.92dB, 0.7567)]
{\includegraphics[width=0.24\linewidth]{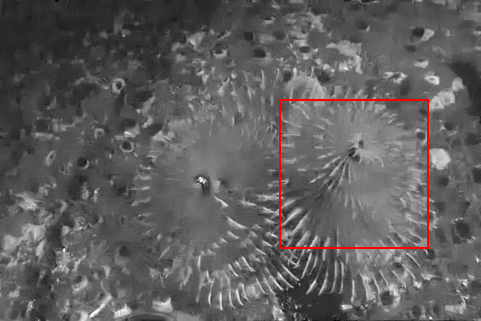}}\hfill
\subfigure[TNRD (28.29dB, 0.7758)]
{\includegraphics[width=0.24\linewidth]{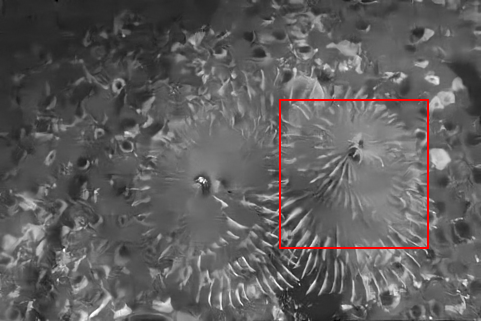}}\hfill
\subfigure[TNLRD (28.39dB, 0.7821, L2 loss)]
{\includegraphics[width=0.24\linewidth]{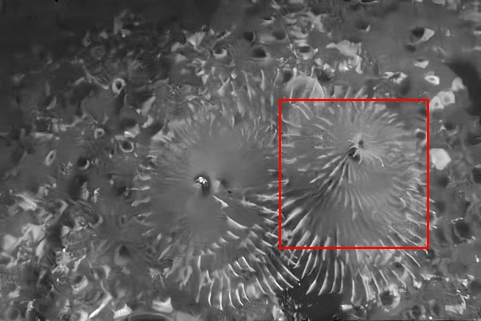}}\hfill
\subfigure[TNLRD (28.17dB, 0.7830, SSIM loss)]
{\includegraphics[width=0.24\linewidth]{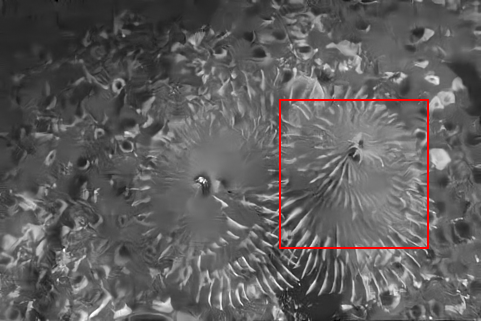}}\\
\vspace{2mm}
\subfigure[original]
{\includegraphics[width=0.24\linewidth]{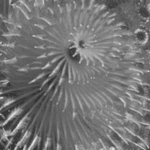}}\hfill
\subfigure[noisy]
{\includegraphics[width=0.24\linewidth]{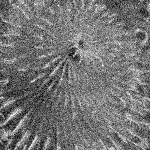}}\hfill
\subfigure[BM3D (27.98dB, 0.7642)]
{\includegraphics[width=0.24\linewidth]{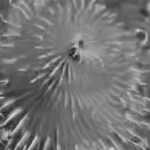}}\hfill
\subfigure[WNNM (28.24dB, 0.7706)]
{\includegraphics[width=0.24\linewidth]{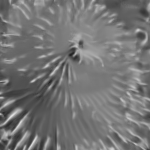}}\\
\vspace{2mm}
\subfigure[NLR-MRF (27.92dB, 0.7567)]
{\includegraphics[width=0.24\linewidth]{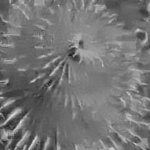}}\hfill
\subfigure[TNRD (28.29dB, 0.7758)]
{\includegraphics[width=0.24\linewidth]{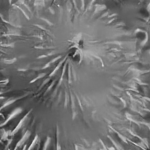}}\hfill
\subfigure[TNLRD (28.39dB, 0.7821, L2 loss)]
{\includegraphics[width=0.24\linewidth]{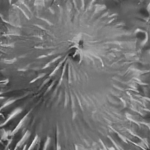}}\hfill
\subfigure[TNLRD (28.17dB, 0.7830, SSIM loss)]
{\includegraphics[width=0.24\linewidth]{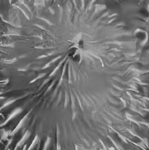}}\\
\caption{Denoising results comparison in 68 test images for $\sigma=25$.
From left to right and from top to down, the images are produced by original, noisy, BM3D, WNNM, NLR-MRF, TNRD, TNLRD trained with L2 loss and TNLRD trained with SSIM loss.
The values in the blankets are PSNR and SSIM values respectively.
The recover comparisons of the texture region are marked with red boxes.}\label{fig:68img12}
\end{figure*}

\begin{figure*}[t!]
\centering
\subfigure[original]
{\includegraphics[width=0.24\linewidth]{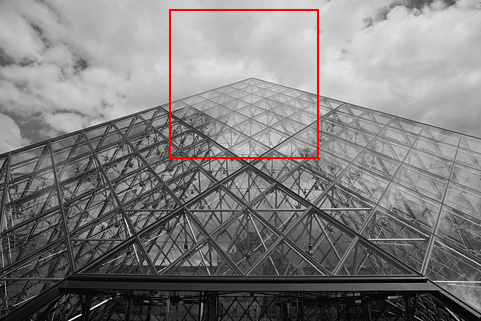}}\hfill
\subfigure[noisy]
{\includegraphics[width=0.24\linewidth]{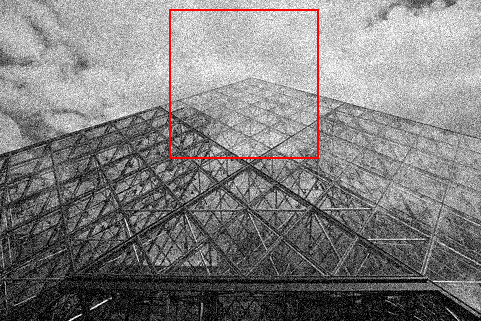}}\hfill
\subfigure[BM3D (27.09dB, 0.8182)]
{\includegraphics[width=0.24\linewidth]{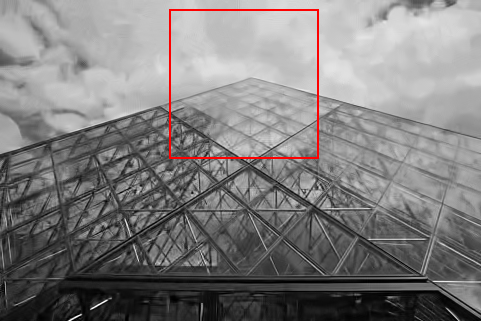}}\hfill
\subfigure[WNNM (27.47dB, 0.8326)]
{\includegraphics[width=0.24\linewidth]{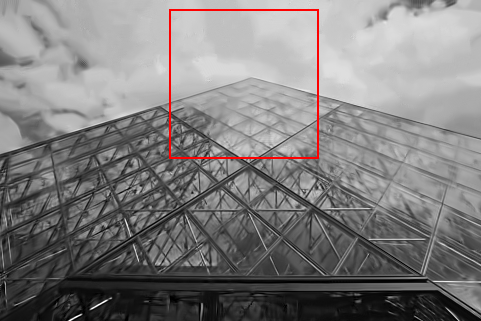}}\\
\vspace{2mm}
\subfigure[NLR-MRF (26.66dB, 0.8015)]
{\includegraphics[width=0.24\linewidth]{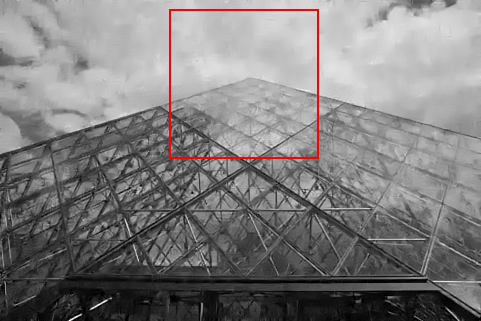}}\hfill
\subfigure[TNRD (27.34dB, 0.8352)]
{\includegraphics[width=0.24\linewidth]{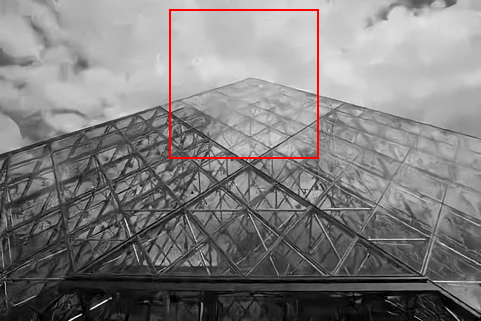}}\hfill
\subfigure[TNLRD (27.61dB, 0.8460, L2 loss)]
{\includegraphics[width=0.24\linewidth]{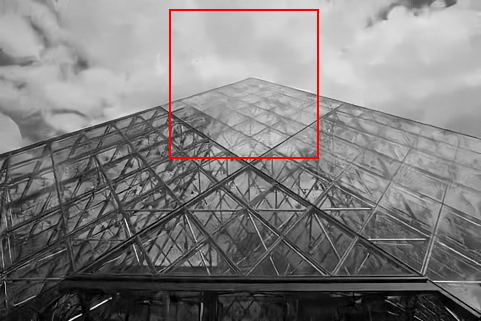}}\hfill
\subfigure[TNLRD (27.35dB, 0.8480, SSIM loss)]
{\includegraphics[width=0.24\linewidth]{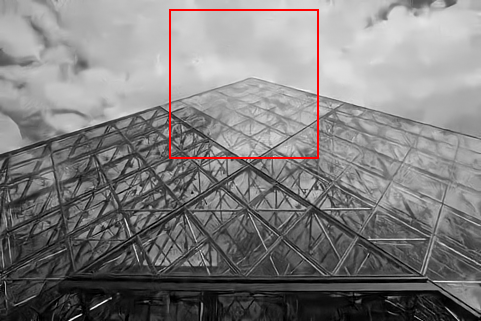}}\\
\vspace{2mm}
\subfigure[original]
{\includegraphics[width=0.24\linewidth]{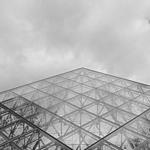}}\hfill
\subfigure[noisy]
{\includegraphics[width=0.24\linewidth]{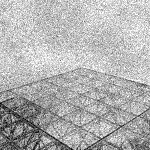}}\hfill
\subfigure[BM3D (27.09dB, 0.8182)]
{\includegraphics[width=0.24\linewidth]{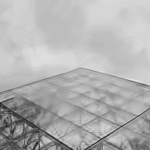}}\hfill
\subfigure[WNNM (27.47dB, 0.8326)]
{\includegraphics[width=0.24\linewidth]{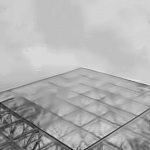}}\\
\vspace{2mm}
\subfigure[NLR-MRF (26.66dB, 0.8015)]
{\includegraphics[width=0.24\linewidth]{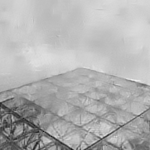}}\hfill
\subfigure[TNRD (27.34dB, 0.8352)]
{\includegraphics[width=0.24\linewidth]{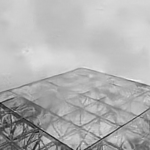}}\hfill
\subfigure[TNLRD (27.61dB, 0.8460, L2 loss)]
{\includegraphics[width=0.24\linewidth]{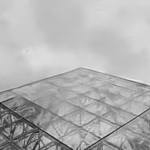}}\hfill
\subfigure[TNLRD (27.35dB, 0.8480, SSIM loss)]
{\includegraphics[width=0.24\linewidth]{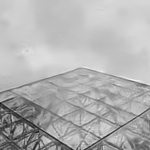}}\\
\caption{Denoising results comparison in 68 test images for $\sigma=25$.
From left to right and from top to down, the images are produced by original, noisy, BM3D, WNNM, NLR-MRF, TNRD, TNLRD trained with L2 loss and TNLRD trained with SSIM loss.
The values in the blankets are PSNR and SSIM values respectively.
The recover comparisons of the texture region are marked with red boxes.}\label{fig:68img44}
\end{figure*}

\begin{table}[t!]
\centering
\begin{threeparttable}
\caption{Average PSNR(dB) on 68 images from \cite{chen2015learning} for image denoising with $\sigma=15$, $25$ and $50$.}\label{table:68imagesPSNR}
\begin{tabular}{|c|c|c|c|c|c|c|}
\hline
\multirow{3}*{$\sigma$} & \multicolumn{6}{c|}{Method} \\
\cline{2-7}
& {BM3D} &{WNNM} &{NLR-MRF} &{TNRD} &\multirow{2}{*}{TNLRD\tnote{*}} &\multirow{2}*{TNLRD\tnote{**}} \\
& {\cite{dabov2007image}} &{\cite{gu2014weighted}} &{\cite{sun2011learning}} &{\cite{chen2015learning}} &{} & {} \\
\hline
15 & 31.08 & 31.37 & 30.97 & 31.42 & \textbf{31.50} & 31.31 \\
25 & 28.56 & 28.83 & 28.48 & 28.91 & \textbf{29.01} & 28.83 \\
50 & 25.62 & 25.83 & 25.38 & 25.96 & \textbf{26.06} & 25.80 \\
\hline
\end{tabular}
\begin{tablenotes}
\item[*] trained with L2 loss.
\item[**] trained with SSIM loss.
\end{tablenotes}
\end{threeparttable}
\end{table}

\begin{table}[t!]
\centering
\begin{threeparttable}
\caption{Average SSIM on 68 images from \cite{chen2015learning} for image denoising with $\sigma=15$, $25$ and $50$.}\label{table:68imagesSSIM}
\begin{tabular}{|c|c|c|c|c|c|c|}
\hline
\multirow{3}*{$\sigma$} & \multicolumn{6}{c|}{Method} \\
\cline{2-7}
& {BM3D} &{WNNM} &{NLR-MRF} &{TNRD} &\multirow{2}{*}{TNLRD\tnote{*}} &\multirow{2}*{TNLRD\tnote{**}} \\
& {\cite{dabov2007image}} &{\cite{gu2014weighted}} &{\cite{sun2011learning}} &{\cite{chen2015learning}} &{} & {} \\
\hline
15 & 0.8717 & 0.8759 & 0.8699 & 0.8821 & 0.8852 & \textbf{0.8864} \\
25 & 0.8013 & 0.8084 & 0.7972 & 0.8152 & 0.8201 & \textbf{0.8219} \\
50 & 0.6864 & 0.6981 & 0.6665 & 0.7024 & 0.7094 & \textbf{0.7113} \\
\hline
\end{tabular}
\begin{tablenotes}
\item[*] trained with L2 loss.
\item[**] trained with SSIM loss.
\end{tablenotes}
\end{threeparttable}
\end{table}

The above training experiments are conducted on Gaussian noise level $\sigma=25$.
We also trained the proposed TNLRD models for the noise level $\sigma=15$ and $\sigma=50$.
After training the models,
we evaluated them on the 68 test images used in \cite{chen2015learning}.
We also tested the TNRD models and our TNLRD models on the 9 test image set.

\begin{figure}[t!]
\centering
\subfigure[The average PSNRs produced by these methods are illustrated in Table \ref{table:68imagesPSNR}.
The markers under the red line means our TNLRD models offer better PSNRs than the comparison methods,
while markers under the red line means the comparison methods offer better PSNRs.
Our TNLRD models outperform BM3D, NLR-MRF and TNRD on all test images,
while WNNM can produce better results on few test images.]
{\includegraphics[width=0.6\linewidth]{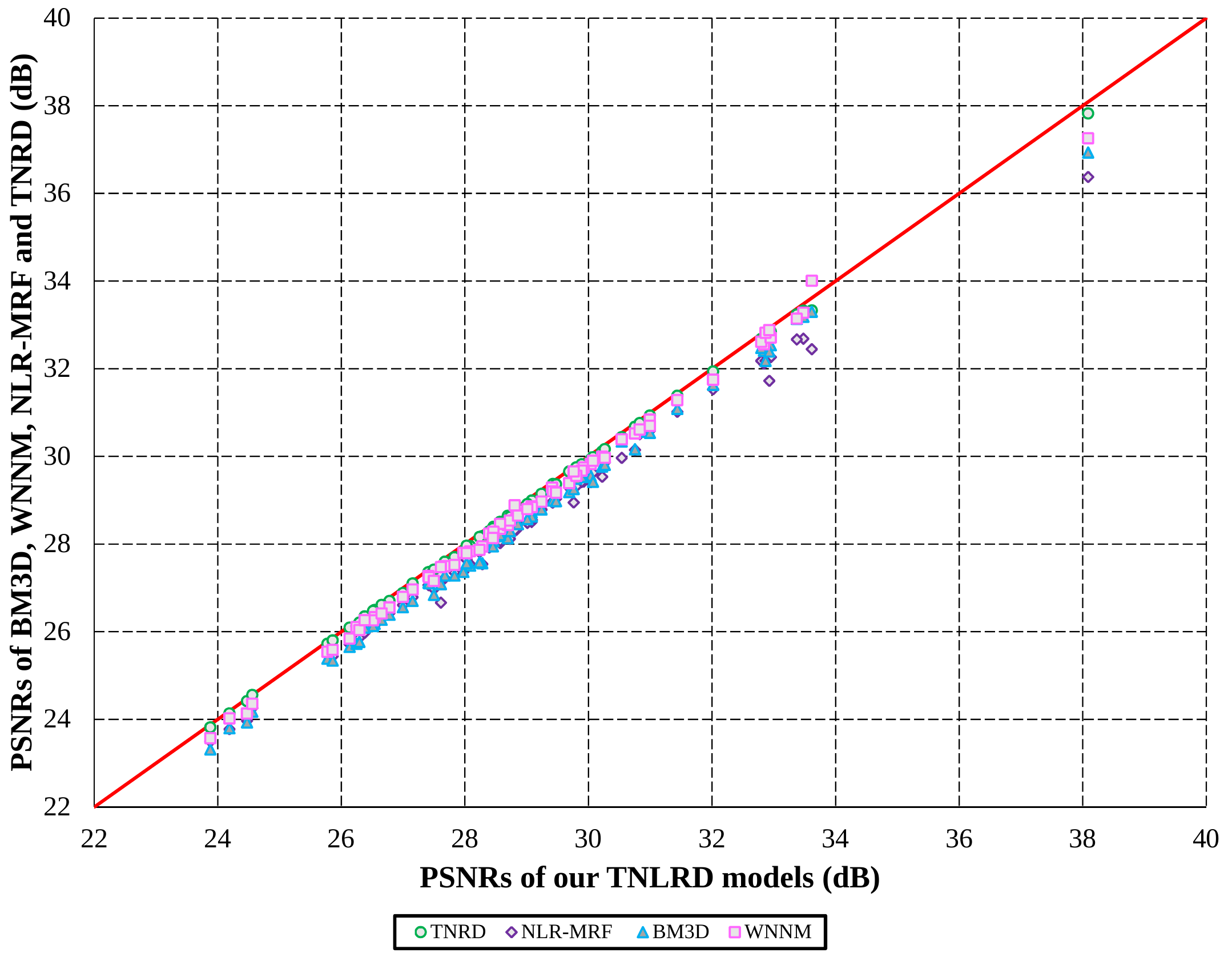}}
\subfigure[The average SSIM values produced by these methods are illustrated in Table \ref{table:68imagesSSIM}.
Our TNLRD models outperform BM3D, NLR-MRF and TNRD on all test images,
while WNNM can produce better results on few test images.]
{\includegraphics[width=0.6\linewidth]{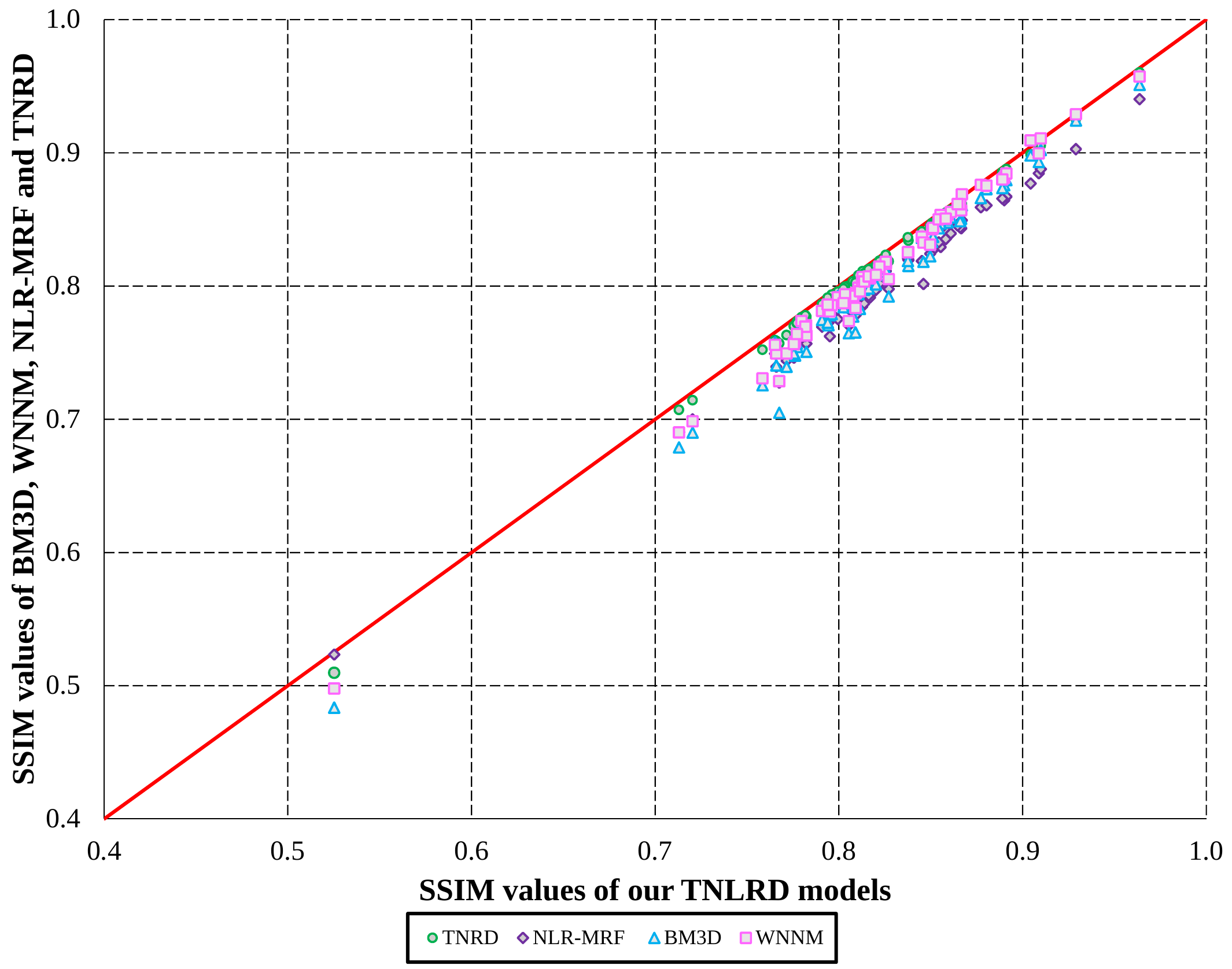}}
\caption{
Scatter plots of the PSNRs and SSIM over 68 test images
with $\sigma = 25$ produced by BM3D, WNNM, NLR-MRF, TNRD and our TNLRD models.
}\label{fig:68imgPSNR}
\end{figure}

The denoising performance on the 68 test images
is summarized in Table \ref{table:68imagesPSNR} and \ref{table:68imagesSSIM},
compared with some recent state-of-the-art denoising algorithms.
As illustrated in Table \ref{table:68imagesPSNR} and Fig. \ref{fig:68imgPSNR},
the proposed TNLRD models outperform
the TNRD models by almost 0.1dB,
BM3D by 0.45dB,
WNNM by 0.18dB and
NLR-MRF by 0.53dB.
In Fig. \ref{fig:68img12} (i-p), we can see that
our TNLRD models recover more clear stems in the sea anemone than the TNRD models.
While BM3D and WNNM tend to over-smooth texture regions,
our TNLRD models produce sharper recovered image.
In Fig. \ref{fig:68img44} (i-p), we can also see that
clear and straight steel structures are recovered by our TNLRD models,
while the TNRD models tends to offer the over-smooth results in the texture regions.
The same phenomenon can be also found in the recovered image produced by BM3D and WNNM.
Taking a close look at the recovered images produced by BM3D, WNNM and TNRD,
one can see some artifacts in the plain regions.

\begin{figure}[t!]
\centering
\subfigure[original]
{\includegraphics[width=0.48\linewidth]{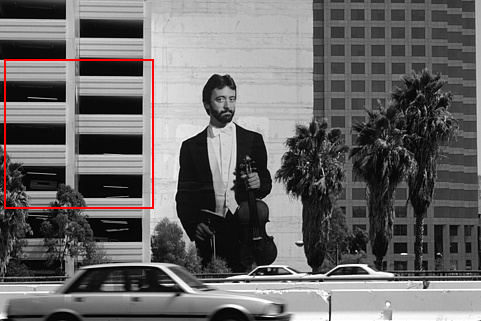}}\hfill
\subfigure[noisy]
{\includegraphics[width=0.48\linewidth]{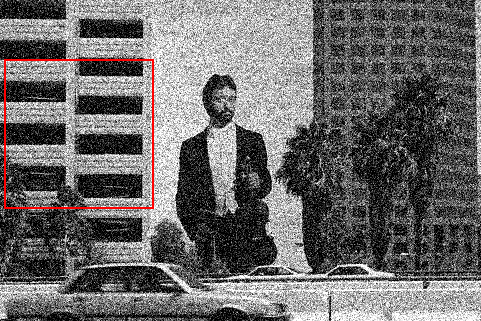}}\\
\subfigure[TNRD (25.05dB, 0.7431)]
{\includegraphics[width=0.48\linewidth]{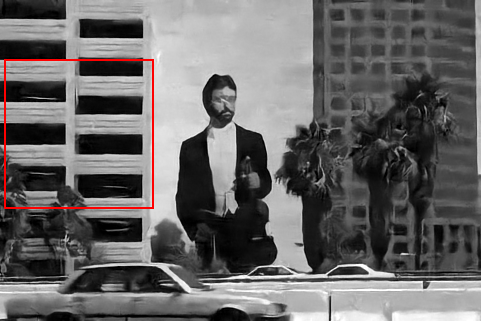}}\hfill
\subfigure[TNLRD (25.25dB, 0.7521)]
{\includegraphics[width=0.48\linewidth]{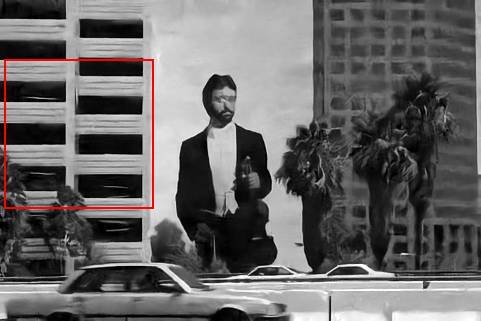}}\\
\vspace{2mm}
\subfigure[original]
{\includegraphics[width=0.2\linewidth]{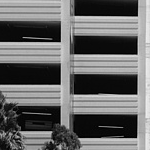}}\hfill
\subfigure[noisy]
{\includegraphics[width=0.2\linewidth]{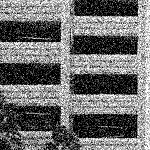}}\hfill
\subfigure[TNRD]
{\includegraphics[width=0.2\linewidth]{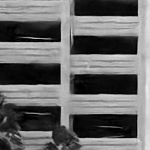}}\hfill
\subfigure[TNLRD]
{\includegraphics[width=0.2\linewidth]{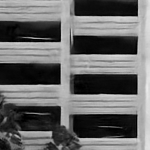}}\\
\caption{Denoising results comparison in 68 test images for $\sigma=50$.
From left to right and from top to down, the images are produced by original, noisy, TNRD and TNLRD.
The values in the blankets are PSNR and SSIM values respectively.}\label{fig:68img11}
\end{figure}
\begin{figure}[t!]
\centering
\subfigure[original]
{\includegraphics[width=0.24\linewidth]{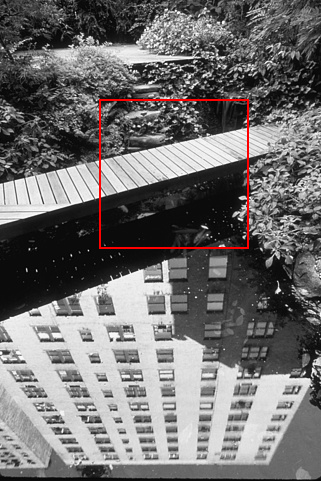}}\hfill
\subfigure[noisy]
{\includegraphics[width=0.24\linewidth]{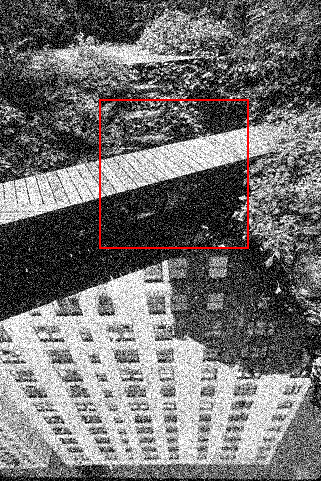}}\hfill
\subfigure[TNRD (22.43dB, 0.7262)]
{\includegraphics[width=0.24\linewidth]{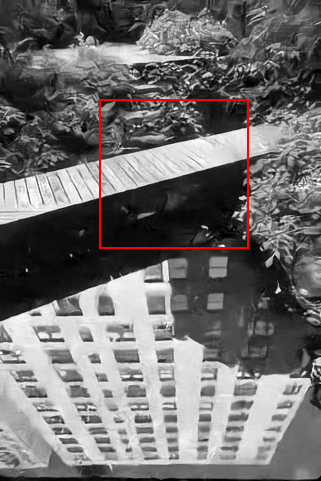}}\hfill
\subfigure[TNLRD (22.61dB, 0.7417)]
{\includegraphics[width=0.24\linewidth]{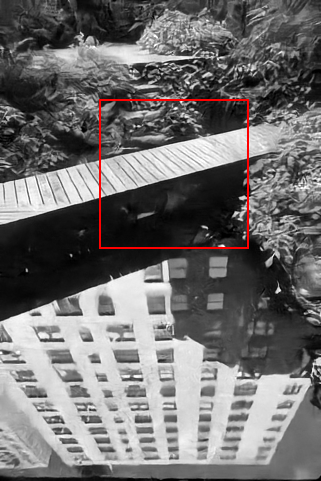}}\\
\vspace{2mm}
\subfigure[original]
{\includegraphics[width=0.2\linewidth]{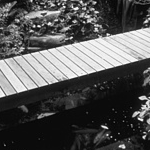}}\hfill
\subfigure[noisy]
{\includegraphics[width=0.2\linewidth]{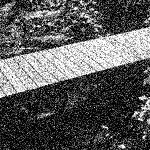}}\hfill
\subfigure[TNRD]
{\includegraphics[width=0.2\linewidth]{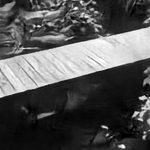}}\hfill
\subfigure[TNLRD]
{\includegraphics[width=0.2\linewidth]{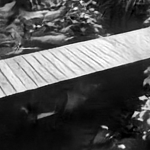}}\\
\caption{Denoising results comparison in 68 test images for $\sigma=50$.
From left to right and from top to down, the images are produced by original, noisy, TNRD and TNLRD.
The values in the blankets are PSNR and SSIM values respectively.}\label{fig:68img21}
\end{figure}

We also compared our TNLRD model with these methods
for cases of $\sigma=15$ and $\sigma=50$, as shown in Table \ref{table:68imagesPSNR}.
When the image is heavily degraded by the noise, i.e., $\sigma$ is getting larger,
the local methods, e.g., the TNRD model,
can not collect enough information for inference,
and may create artifacts and remove textures.
On the contrary, the non-local methods collect more information,
and tackle the artifacts and preserve textures.
We show some denoising examples with $\sigma=50$
in Fig. \ref{fig:68img11} and \ref{fig:68img21}.

\begin{figure}[t!]
\centering
\subfigure[original]
{\includegraphics[width=0.48\linewidth]{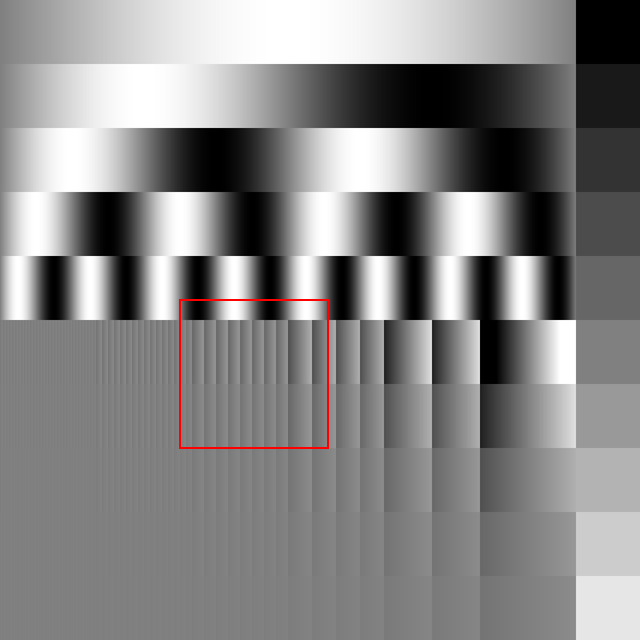}}\hfill
\subfigure[noisy]
{\includegraphics[width=0.48\linewidth]{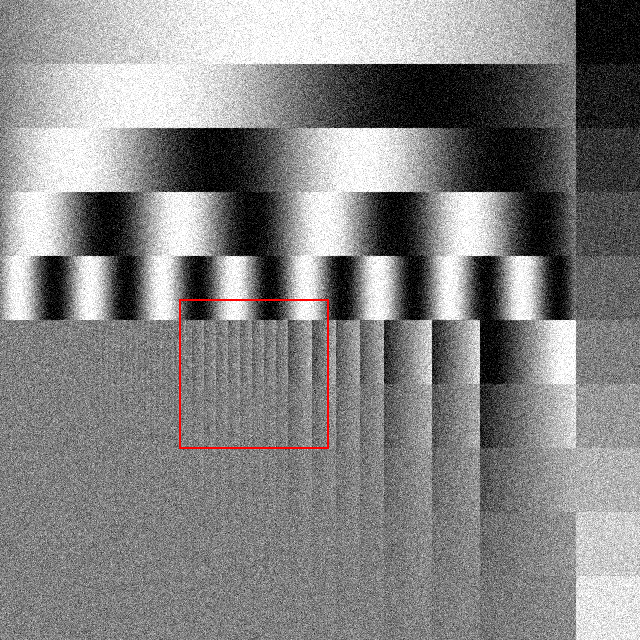}}\\
\subfigure[TNRD (37.82dB)]
{\includegraphics[width=0.48\linewidth]{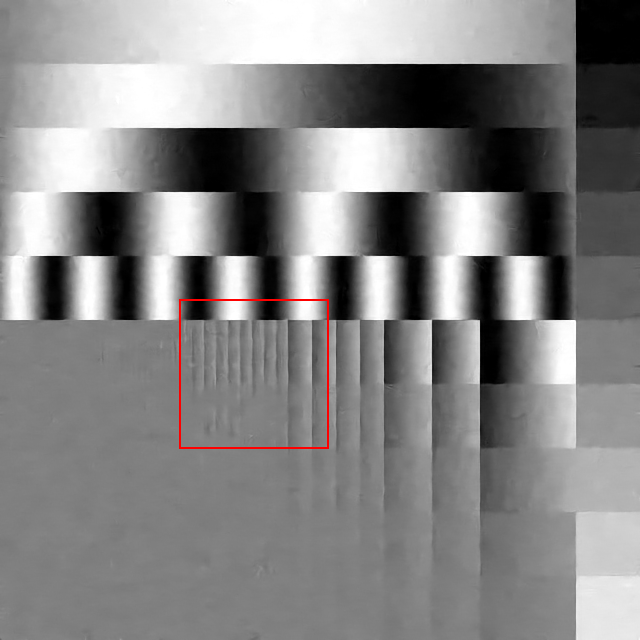}}\hfill
\subfigure[TNLRD (38.41dB)]
{\includegraphics[width=0.48\linewidth]{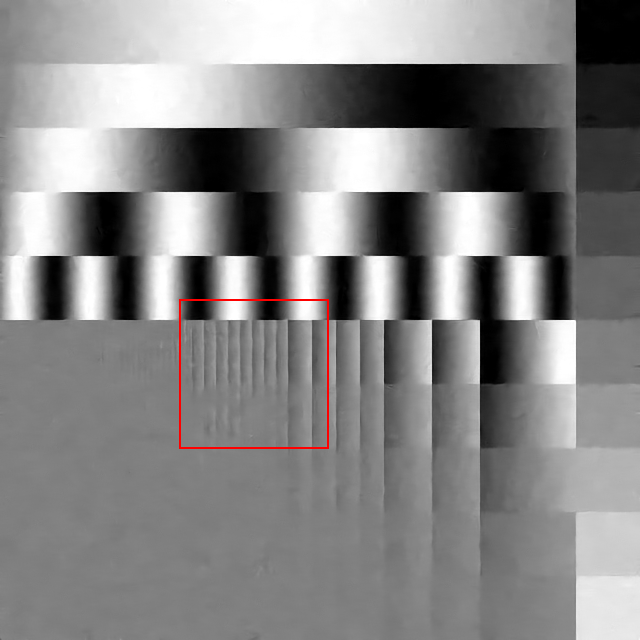}}\\
\vspace{2mm}
\subfigure[original]
{\includegraphics[width=0.2\linewidth]{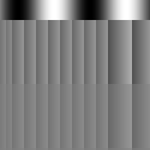}}\hfill
\subfigure[noisy]
{\includegraphics[width=0.2\linewidth]{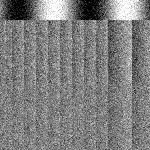}}\hfill
\subfigure[TNRD]
{\includegraphics[width=0.2\linewidth]{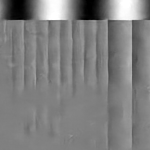}}\hfill
\subfigure[TNLRD]
{\includegraphics[width=0.2\linewidth]{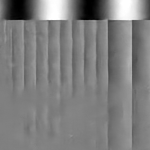}}\\
\caption{Denoising results comparison of $2^{th}$ in 9 test images.
From left to right and from top to down,
the images are produced by original, noisy, TNRD and TNLRD.}\label{fig:9img1}
\end{figure}

\begin{figure}[t!]
\centering
\subfigure[original]
{\includegraphics[width=0.48\linewidth]{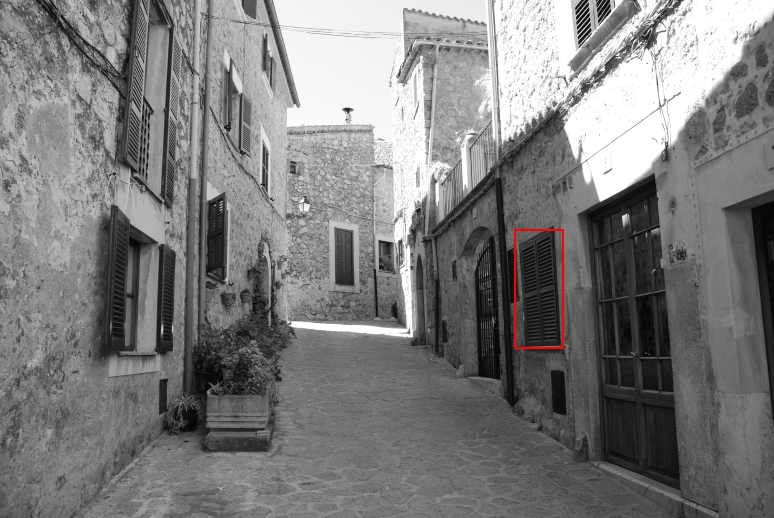}}\hfill
\subfigure[noisy]
{\includegraphics[width=0.48\linewidth]{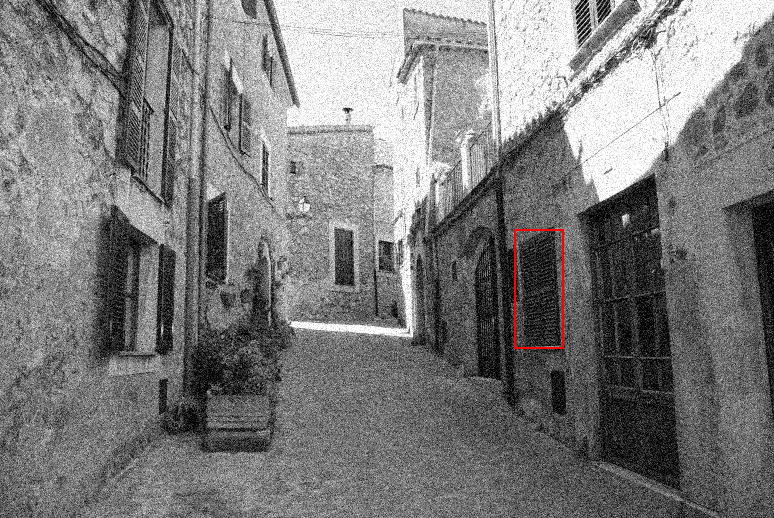}}\\
\subfigure[TNRD (28.61dB)]
{\includegraphics[width=0.48\linewidth]{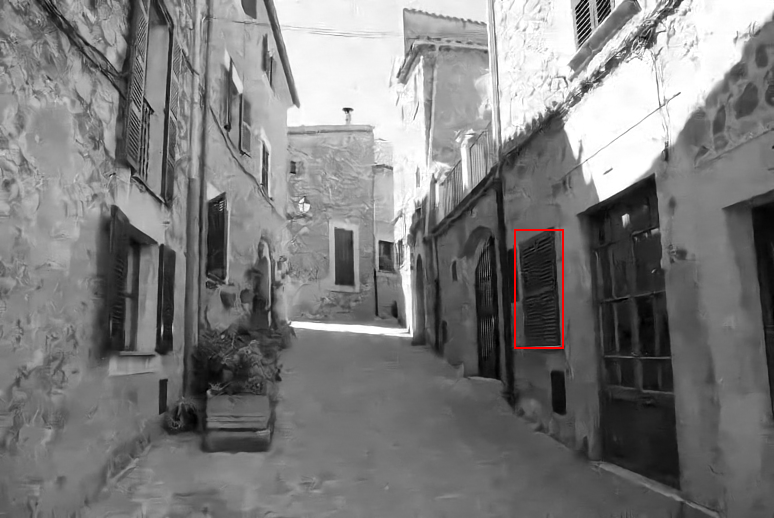}}\hfill
\subfigure[TNLRD (28.77dB)]
{\includegraphics[width=0.48\linewidth]{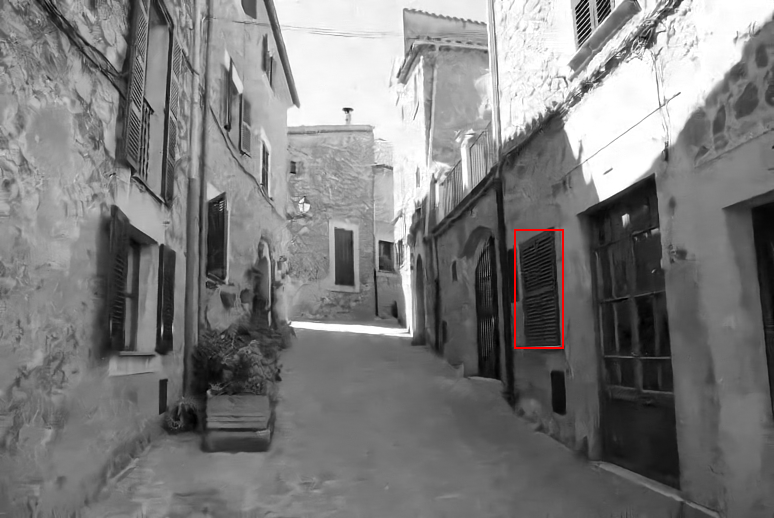}}\\
\vspace{2mm}
\subfigure[original]
{\includegraphics[width=0.12\linewidth]{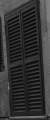}}\hfill
\subfigure[noisy]
{\includegraphics[width=0.12\linewidth]{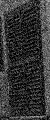}}\hfill
\subfigure[TNRD]
{\includegraphics[width=0.12\linewidth]{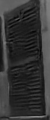}}\hfill
\subfigure[TNLRD]
{\includegraphics[width=0.12\linewidth]{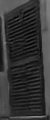}}\\
\caption{Denoising results comparison of $3^{th}$ in 9 test images.
From left to right and from top to down,
the images are produced by original, noisy, TNRD and TNLRD.}\label{fig:9img2}
\end{figure}

\begin{figure}[t!]
\centering
\includegraphics[width=0.9\linewidth]{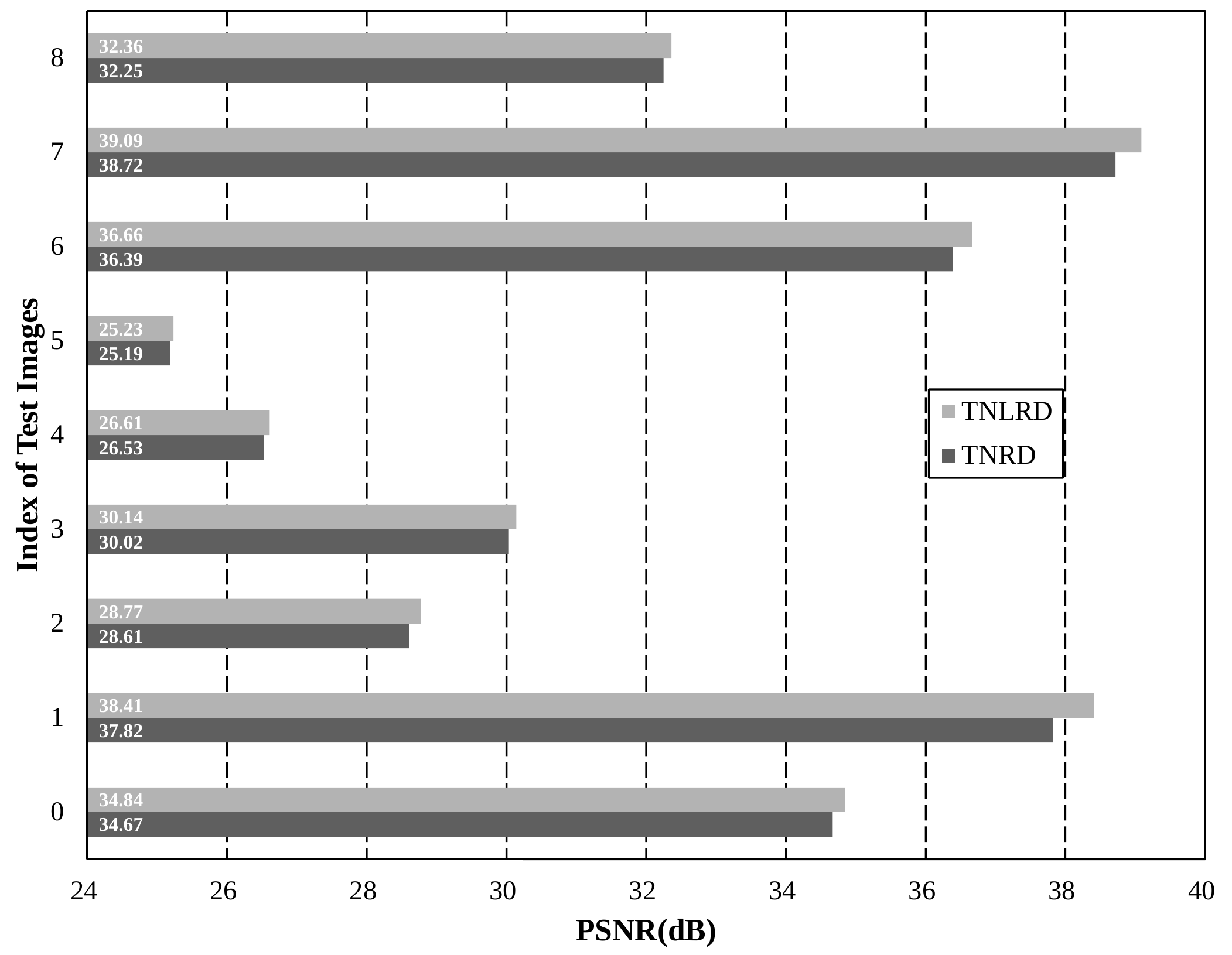}
\caption{Bar plots of PSNRs over 9 test images with $\sigma=25$ produced by TNRD and our TNLRD models.
The bars in red donate the PSNRs produced by our TNLRD models,
the bars in blue are for TNRD models.}\label{fig:9imgPSNR}
\end{figure}

We also compare our TNLRD and TNRD on the 9 test images collected from web.
In Fig. \ref{fig:9img1} (e-h), we can see that
our TNLRD models recover the vertical lines more clear than the TNRD models.
In Fig. \ref{fig:9img2} (e-h), we can see that
our TNLRD models recover the window structures more precisely than the TNRD models.
In Fig. \ref{fig:9imgPSNR}, we can conclude that our TNLRD models
surpass the TNRD models for each test image.
The average PSNR produced by our TNLRD and TNRD models are 32.46dB and 32.24dB respectively.

From the detailed comparison with some state-of-the-art denoising methods,
especially the newly proposed TNRD,
we can conclude that our TNLRD models offer better quality and quantity performance in Gaussian denoising.

\section{Conclusion}
In this paper, we propose trainable non-local reaction diffusion models for image denoising.
We introduce the NSS prior as non-local filters to the TNRD models.
We train the models parameters, i.e., local linear filters,
non-local filters and non-linear influence functions, in a loss-based learning scheme.
From the comparison with the state-of-the-art image denoising methods,
we concluded that our TNLRD models achieve superior
image denoising performance in terms of both PSNR and SSIM.
Our TNLRD models also provide visually plausible
denoised image with less artifacts and more textures.

\bibliographystyle{IEEEtran}
\bibliography{interpolation}

\end{document}